\documentclass[10pt,twocolumn,letterpaper]{article}

\usepackage[pagenumbers]{iccv} 
\usepackage{multirow,diagbox,makecell,amssymb,bm,bbding}
\usepackage{amsmath,amsthm}
\usepackage{todonotes}
\usepackage{graphicx}
\usepackage{pifont}
\usepackage{algorithm}
\usepackage{algorithmic}
\usepackage{amsfonts}       
\usepackage{booktabs}       
\usepackage{footmisc}
\usepackage{listings}
\usepackage{tcolorbox}
\usepackage{subcaption}
\usepackage{placeins}

\definecolor{iccvblue}{rgb}{0.21,0.49,0.74}
\definecolor{cyanblue}{rgb}{0.39, 0.58, 0.93}
\definecolor{pureblue}{rgb}{0, 1, 1}
\definecolor{tinyred}{rgb}{1, 0.42, 0.42}
\usepackage[pagebackref,breaklinks,colorlinks,allcolors=iccvblue]{hyperref}

\def\paperID{10671} 
\def\confName{ICCV}
\def\confYear{2025}

\title{GEMeX: A Large-Scale, Groundable, and Explainable Medical VQA Benchmark for Chest X-ray Diagnosis}
\newcommand{\shortname}{\text{GEMeX}}

\author{Bo Liu$^{1}$,  Ke Zou$^{2,3}$, Liming Zhan$^{1}$, Zexin Lu$^{1}$, Xiaoyu Dong$^{1}$, Yidi Chen$^{4}$, Chengqiang Xie$^{1}$, \\Jiannong Cao$^{1}$, Xiao-Ming Wu$^{1}$\thanks{~ Corresponding author.}, Huazhu Fu$^{5*}$\\
\small{$^{1}$The Hong Kong Polytechnic University, Hong Kong,}  \small{$^{2}$National University of Singapore, Singapore,} \\
\small{$^{3}$Sichuan University, China,}
\small{$^{4}$West China Hospital of Sichuan University, China}\\
\small{$^{5}$Institute of High Performance Computing, Agency for Science, Technology and Research, Singapore}\\
}

\begin{document}
\maketitle
\begin{abstract}
Medical Visual Question Answering (Med-VQA) combines computer vision and natural language processing to automatically answer clinical inquiries about medical images.
However, current Med-VQA datasets exhibit two significant limitations: (1) they often lack visual and textual explanations for answers, hindering comprehension for patients and junior doctors; (2) they typically offer a narrow range of question formats, inadequately reflecting the diverse requirements in practical scenarios. These limitations pose significant challenges to the development of a reliable and user-friendly Med-VQA system.
To address these challenges, we introduce a large-scale, \textbf{G}roundable, and \textbf{E}xplainable \textbf{Me}dical VQA benchmark for chest \textbf{X}-ray diagnosis (\shortname{}), featuring several innovative components: 
(1) a multi-modal explainability mechanism that offers detailed visual and textual explanations for each question-answer pair, thereby enhancing answer comprehensibility; (2) four question types—open-ended, closed-ended, single-choice, and multiple-choice—to better reflect practical needs. 
With 151,025 images and 1,605,575 questions, \shortname{} is the currently largest chest X-ray VQA dataset.
Evaluation of 12 representative large vision language models (LVLMs) on \shortname{} reveals suboptimal performance, underscoring the dataset's complexity. 
Meanwhile, we propose a strong model by fine-tuning an existing LVLM on the \shortname{} training set. The substantial performance improvement showcases the dataset's effectiveness. The benchmark is available at \url{www.med-vqa.com/GEMeX}.
\end{abstract}    
\begin{figure}[!t]
  \centering
  \includegraphics[width=0.95\linewidth]{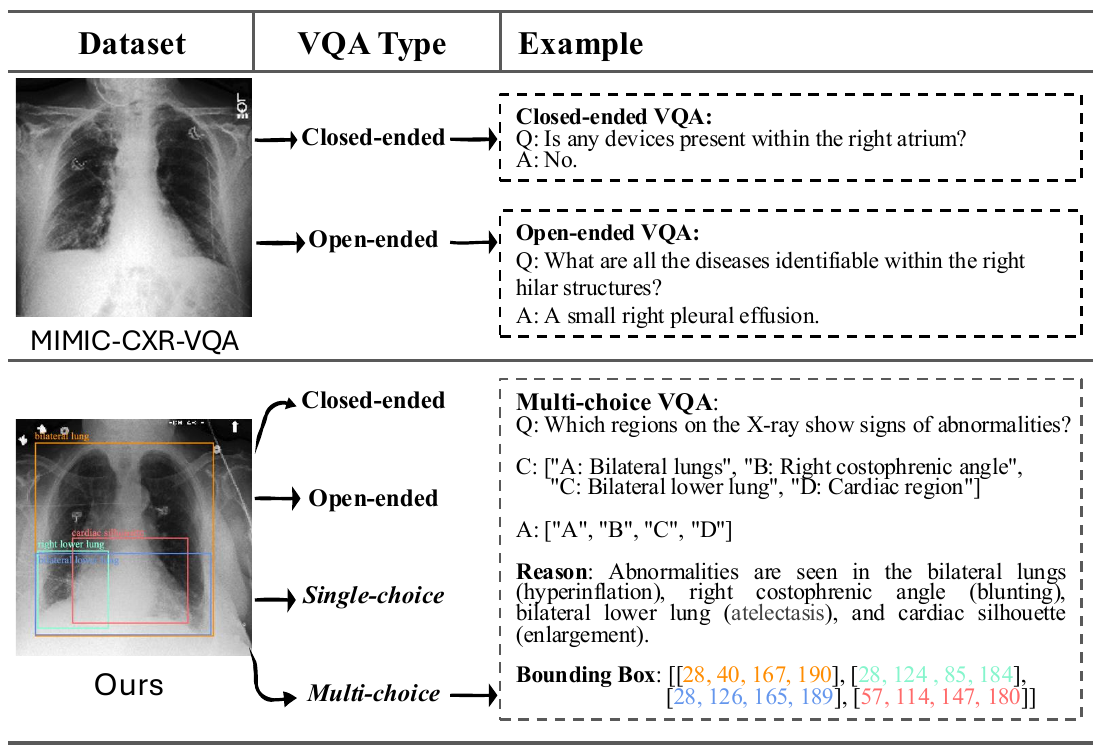}
  \caption{
  Our \shortname{} stands out from existing medical VQA datasets by providing diverse question types and comprehensive multimodal explanations: textual reasoning and visual grounding.
  }
  \label{fig:slake_cxr_feature}
\end{figure}

\begin{table*}[t]
    \centering
    \small
    \resizebox{0.93\linewidth}{!}{
    \begin{tabular}{cccccccc}
    \hline
    \hline
    Dataset & \# Images & \# QA Pairs & \# Modalities & \# Question Types\ddag & \# Groundable  & \# Explainable\\
    \hline
    VQA-RAD~\cite{lau2018dataset} & 0.315K & 3.5K & Diverse\dag & O.  \& C.  &\XSolidBrush & \XSolidBrush\\
    SLAKE~\cite{liu2021slake} & 0.642K & 14K & Diverse\dag&  O.  \& C. & \XSolidBrush & \XSolidBrush\\
    OmniMedVQA~\cite{hu2024omnimedvqa} & 118.010K & 128K &Diverse\dag & O.  \& C. \& S. & \XSolidBrush & \XSolidBrush\\
    PMC-VQA~\cite{zhang2023pmc} & 149.075K & 227K &Diverse\dag&  O.  \& C. \& S. &\XSolidBrush & \XSolidBrush\\
    VQA-Med~\cite{ben2021overview} & 4.5K & 4.5K &Diverse\dag&  O.  \& C. &\XSolidBrush & \XSolidBrush\\
    PathVQA~\cite{he2020pathvqa} & 149K & 33K &Pathology&  O.  \& C. &\XSolidBrush & \XSolidBrush\\
    \hline
    RadGenome-Chest CT~\cite{zhang2024radgenome} & 50.188K & 1.3M & Chest CT& O.  \& C. &\Checkmark & \XSolidBrush\\
    MIMIC-Diff-VQA~\cite{hu2023expert} &164.324K & 700K &Chest X-ray &  O.  \& C.&\XSolidBrush & \XSolidBrush\\
    MIMIC-CXR-VQA~\cite{bae2024ehrxqa} & 142.797K & 377K  &Chest X-ray & O.  \& C. & \XSolidBrush & \XSolidBrush\\
    \textbf{\shortname{} (Ours)} & 151.025K & \textbf{1.6M} &Chest X-ray& O.  \& C. \& S. \& \textbf{M.}&  \Checkmark & \Checkmark (Vision \& Language)\\
    \hline
    \hline
    \end{tabular}
    }
    \caption{
    Comparison of medical VQA Datasets. \dag\ indicates a composition of multiple body parts (e.g., head, chest, abdomen) and various imaging modalities (e.g., CT, MRI, X-ray, pathology). In the \# Question Types\ddag\ column, O., C., S., and M. represent ``Open-ended'', ``Closed-ended'', ``Single-choice'', and ``Multi-choice'', respectively.}  
    
    \label{tab:data_comparison}
\end{table*}

\section{Introduction}
\label{sec:intro}
Large vision language models (LVLMs) have recently made huge advancements in artificial intelligence~\cite{achiam2023gpt, zhao2024easygen,liu2024visual, zhu2023minigpt,alayrac2022flamingo,ramesh2021zero,dai2023instructblip}, demonstrating remarkable capabilities in understanding visual content while generating coherent natural language responses.  
These advancements have driven innovations across various domains~\cite{xu2023lvlm,gu2024anomalygpt}, with healthcare emerging as an important application.
Within this domain, medical visual question answering (Med-VQA) stands out as a crucial task that automatically provides reliable and user-friendly answers~\cite{lin2023medical} to questions about medical images~\cite{lau2018dataset}, facilitating healthcare professionals in diagnosis, medical education, and clinical decision-making.

To ensure the reliability and user-friendliness of Med-VQA systems, it is crucial to incorporate answer explanations along with a diverse set of question formats. Although significant progress has been made by existing Med-VQA systems~\cite{lau2018dataset,liu2021slake,hu2024omnimedvqa,zhang2023pmc, he2020pathvqa}, none have yet integrated explanations for the answers, especially in terms of the visual aspect. As emphasized by \cite{li2018vqa}, explanations are as essential as the answers themselves in general VQA systems. This holds even stronger in medical VQA, where the domain-specific nature of the task amplifies the need for clarity~\cite{lau2018dataset, zhao2024chatcad}. 
Additionally, the limited range of question formats, such as the absence of multiple-choice questions, restricts the real-world applicability of medical AI systems.

To tackle the aforementioned limitations, we develop a large-scale, \textbf{G}roundable, and \textbf{E}xplainable \textbf{Me}dical VQA benchmark for chest \textbf{X}-ray diagnosis (\shortname{}). 
We first undertake a comprehensive data refinement process upon the Chest ImaGenome~\cite{wu2021chest}. By collaborating with radiologists, we systematically redefine anatomical regions and establish more precise vision-text correspondence mappings, resulting in accurate region-grounded reports for each X-ray image.
Subsequently, we leverage GPT-4o~\cite{achiam2023gpt} to generate a diverse set of questions based on these grounded reports, covering four categories of varying difficulty levels: open-ended, closed-ended, single-choice, and multiple-choice questions. 
Each question-answer pair is enriched with explicit reasoning and corresponding visual region annotations, as shown in Figure~\ref{fig:slake_cxr_feature}.
The resulting dataset comprises 151,025 images and 1,605,575 questions. \emph{Currently, this is the largest VQA dataset for chest X-rays and the first medical VQA dataset that simultaneously includes both textual and visual explanations}.


We evaluate 12 representative LVLMs, including 7 from the general domain (\eg, LLaVA~\cite{liu2024visual}, DeepSeek-VL~\cite{lu2024deepseek}, GPT-4o-mini~\cite{achiam2023gpt}), and 5 from the medical domain (\eg, LLaVA-Med~\cite{li2024llava}, XrayGPT~\cite{thawkar2023xraygpt}, RadFM~\cite{wu2023towards}). The experimental findings underscore the challenging characteristics of our dataset. Additionally, we propose a simple instruction-tuning strategy that derives a task-specific LVLM. The impressive performance improvement highlights the effectiveness of our dataset. Overall, we develop three metrics for measuring the accuracy of model outputs in terms of answers, reasoning, and visual grounding (localization generation). Notably, we apply both semantics-level score and gram-based metrics of natural language generation (\eg, BLEU and ROUGE) for textual parts. Results indicate that for models without \shortname{} fine-tuning, semantics-level scoring is more reliable. After fine-tuning, however, the natural language generation metrics can better reflect the model's understanding of the dataset. 

This manuscript makes the following key contributions:
\begin{itemize}
    \item We present \shortname{}, a large-scale medical VQA dataset for chest X-rays, designed to support diverse question types and provide enhanced explainability for medical VQA systems. To our knowledge, it is the largest chest X-ray VQA dataset and the first Med-VQA dataset to embody the concept of multimodal explainability.
    
    \item We systematically benchmark $12$ representative LVLMs using \shortname{}, introducing multiple evaluation metrics to comprehensively demonstrate the performance of current popular LVLMs on the Med-VQA task.

    %

    
    

    \item Our method shows that integrating precise vision-text explainability enhances the visual reasoning of LVLMs, addressing a key limitation in many models. We emphasize the need for a large-scale, groundable, and explainable VQA benchmark to advance LVLM development and deployment in healthcare.
    
    
    

\end{itemize}

\begin{figure*}[t]
    \centering
    \includegraphics[width=1.0\linewidth]{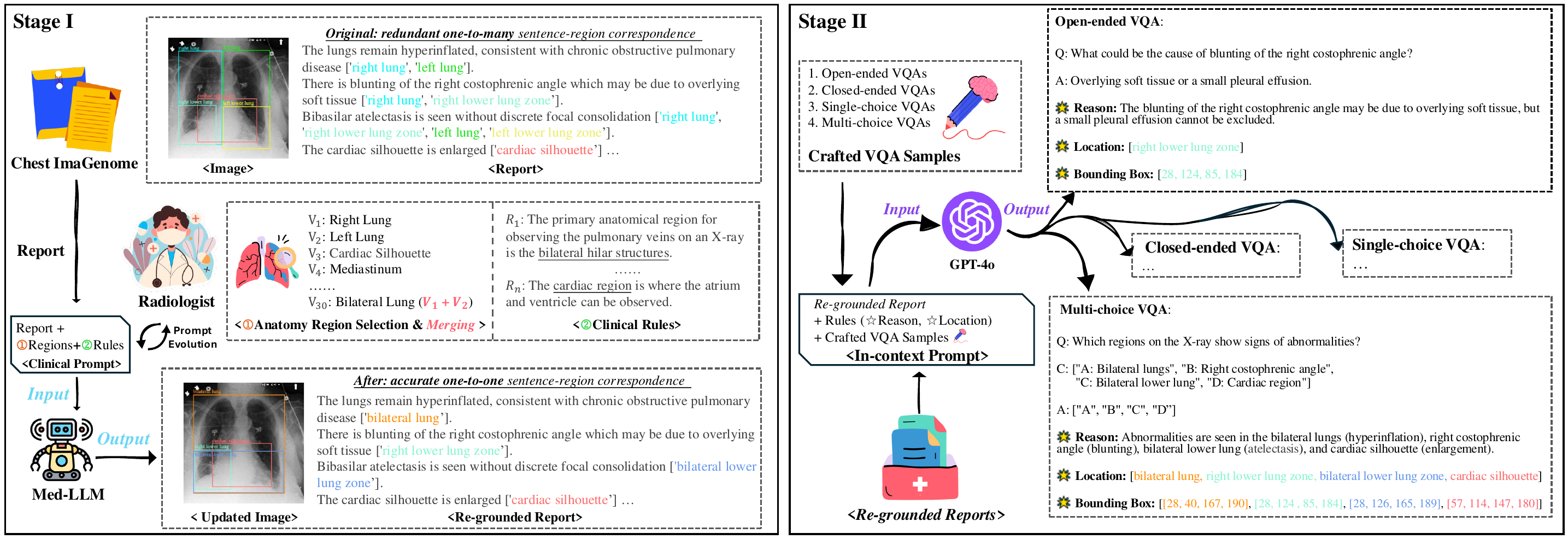}
    \caption{
    Illustration of the proposed pipeline for constructing our \shortname{}, with two main stages. 
    In Stage I (left), medical LLM performs re-grounding on the original reports based on the pathological regions and clinical guidance specified by the radiologists, generating more precise sentence-region correspondence.
    In Stage II (right), the well-crafted prompt enables GPT-4o to generate a high-quality, large-scale Med-VQA dataset with both textual and visual explanations, leveraging the re-grounded reports from Stage I.
    }
    \label{fig:data_construction}
\end{figure*}

\section{Related Work}
\label{sec:related}

\subsection{Medical VQA Datasets}
In recent years, various datasets have been created to advance medical VQA research, each tackling specific challenges in clinical domains. A detailed comparison to other VQA datasets is in Table~\ref{tab:data_comparison}. Specifically,
VQA-RAD~\cite{lau2018dataset} is a pioneer dataset that offers over 3,000 question-answer pairs focused on radiology images.
SLAKE~\cite{liu2021slake} is the first manually created dataset with over 14,000 QA pairs across CT, MRI, and X-ray images, enabling models to handle complex scenarios by combining visual and textual information. 
VQA-Med~\cite{ben2021overview} is a key dataset for Med-VQA competitions, with 4,500 radiology images and paired QAs. 
OmniMedVQA~\cite{hu2024omnimedvqa} provides more data and imaging modalities, which cover the entire body, to encourage model generalization. 
PMC-VQA~\cite{zhang2023pmc} generates VQA data by prompting a large language model to decompose captions of biomedical figures, enabling academic knowledge extraction. 
PathVQA~\cite{he2020pathvqa} supplies over 32,000 QA pairs on histopathological images for fine-grained analysis.  
 

For specialized tasks, RadGenome-Chest CT~\cite{zhang2024radgenome} supports chest CT diagnostics, while MIMIC-Diff-VQA~\cite{hu2023expert} emphasizes differential diagnosis reasoning between two X-rays. MIMIC-CXR-VQA~\cite{bae2024ehrxqa} expands MIMIC-CXR~\cite{johnson2019mimic} with diverse question templates to generate thoracic radiology QA pairs, aiding in chest abnormality detection. However, all current datasets lack explainability and diverse question formats. They do not provide detailed visual and textual explanations for answers, which limits their usability for patient and junior doctor comprehension. Additionally, limited question types restrict their ability to simulate the variety of inquiries encountered in practice.

\subsection{Medical VQA Methods}
Inspired by the advancements in general VQA, medical VQA has gained significant traction as a specialized domain. 
Due to data limitations, however, most approaches~\cite{lau2018dataset,liu2022medical,gong2021cross, khare2021mmbert, moon2022multi,xu2023multi} have focused on directly embedding visual and textual information jointly to capture their relationships. 
With the rise of contrastive language-image pertaining (CLIP)~\cite{radford2021learning}, methods~\cite{chen2022multi,zhang2023biomedclip, chen2023towards} start to focus on applying CLIP to Med-VQA.
A promising way is to fine-tune CLIP’s joint embeddings to better handle specific medical domains, enhancing the model's understanding of clinical questions and visual features~\cite{lin2023pmc}. 
Recently, the explosion of large vision language models has further pushed the boundaries of medical domain~\cite{thawkar2023xraygpt,wu2023towards,alkhaldi2024minigpt,li2024llava,zou2024medrg,lee2023llm}. Generally, they first pre-train models on a large-scale image-text dataset (like PMC-OA~\cite{lin2023pmc}, PMC-15M~\cite{zhang2023biomedclip}) to map visual features into language model's embedding space and then further tune with instruction data for medical consultation~\cite{liu2024visual,alkhaldi2024minigpt} or disease diagnosis~\cite{zhou2023skingpt,thawkar2023xraygpt,chen2024eyegpt,wang2024pathology}. 
These models are now leveraged for Med-VQA tasks to provide richer, more context-aware answers, extending beyond simple text-image alignment to incorporate broader knowledge-based reasoning. 
However, current methods lack explainability, especially in the visual aspect, due to the corresponding absence in current Med-VQA datasets.

\section{Construction of \shortname{} }
 We propose a two-stage pipeline for constructing our \emph{\shortname{}} dataset, accompanied by an overview in Figure~\ref{fig:data_construction}.

\subsection{Re-grounding Reports}
\label{sec:step_1}
As shown in stage I of Figure~\ref{fig:data_construction}, we build upon Chest ImaGenome~\cite{wu2021chest} to construct our dataset, but with an emphasis on the mapping precision between visual regions and textually described entities.
After consulting radiologists, we find that the anatomical descriptions of regions of interest in ImaGenome are imprecise and redundant, introducing ambiguity into clinical diagnoses.
Specifically, a single sentence can be associated with multiple anatomical regions, \eg, ``There is blunting of the right costophrenic angle which may be due to overlying soft tissue'' corresponding to ``right lung'' and ``right lower lung zone''. This redundancy poses challenges in training models to precisely visual grounding. Hence, we perform re-grounding to ensure each sentence is linked to a single, precise region.


\textbf{Anatomical Region Selection and Merging.}\label{sec:region_selection_merge}
In the original Chest ImaGenome, there are 29 significant pathological regions (with bounding boxes). However, in alignment with radiologists' practices, our dataset focuses on retaining core regions that are crucial for diagnosing diseases through X-rays, such as the ``left lower lung'' and ``mediastinum''. Less significant or marginal areas are excluded to streamline the diagnostic training process and enhance clinical relevance, like ``carina'' which is not considered a core region, and ``clavicle'' which accounts for only about $2\%$ of the total regional frequency. 
Furthermore, to enhance clarity and ensure that each sentence corresponds to a single pathological region with finer granularity, semantically similar regions are merged. 
For instance, the ``left lower lung zone'' and ``right lower lung zone'' are combined into a ``bilateral lower lung zone''. This aligns with conditions like ``bibasilar atelectasis'', as illustrated in Stage I of Figure~\ref{fig:data_construction}, where the condition is described as ``Bibasilar atelectasis is seen without discrete focal consolidation''. In total, we define 30 anatomical regions. Detailed transformation from Chest ImaGenome to ours can be found in the Appendix.


\textbf{Report Re-grounding Using Medical LLM.}\label{sec:report_refinement} Then, we utilize OpenBioLLM-70B\footnote{https://huggingface.co/aaditya/Llama3-OpenBioLLM-70B}, known for its outstanding performance across various medical NLP tasks, to \emph{re-ground reports by re-selecting a region for each sentence}. 
To test the effectiveness of the prompt, we begin by randomly selecting 100 pairs from the Chest ImaGenome test set, which includes approximately 367 sentences. 
Initially, the performance of the LLM is suboptimal due to: 
(1) inner knowledge about X-ray disease observation areas is not sufficiently precise, as OpenBioLLM is an NLP model that lacks clinical expertise, and (2) when a sentence indeed involves multiple regions that cannot be merged, the model may either output multiple regions or arbitrarily select one. 

\emph{Iterative Prompt Refinement via Radiologist Feedback:} To address these limitations, we employ an iterative approach, gradually incorporating clinical guidance from radiologists and manually-labeled such cases (in-context learning) to refine the prompt, facilitating (1) more accurate region selection; and (2) more effective sentence splitting and rewriting. 
For example, ``The cardiomediastinal silhouette is normal.'' is converted into \{``The cardiac silhouette is normal.'':``cardiac silhouette'', ``The mediastinal silhouette is normal.'':``mediastinum''\}, where ``cardiomediastinal'' corresponds to the ``cardiac silhouette'' and ``mediastinum''.
This approach ensures the output clauses align one-to-one with the respective regions. Ultimately, the final prompt is determined with an accuracy of approximately $98.4\%$ on the aforementioned test set, provided in the Appendix.

 Figure~\ref{fig:data_construction} presents an example of a re-grounded report and its corresponding updated image.

\subsection{Groundable and Explainable VQA Generation}
\label{sec:step_2}

Although there are many Med-VQA datasets~\cite{hu2023expert, bae2024ehrxqa} available, some even generated using MIMIC-CXR or Chest ImaGenome, they all have two weaknesses that diminish their practicality: (1) lacking strong explainability, especially the visual guidance, that hinders the user's understanding;
 (2) a restricted range of question types, typically confined to open-ended or closed-ended formats with no inclusion of choice-based questions, reducing the flexibility and comprehensiveness.
In general, these issues highlight the necessity for more versatile and explainable Med-VQA datasets to enhance their utility in clinical settings.

\textbf{Data Generation with Quality Control.}
As shown in Stage II of Figure~\ref{fig:data_construction}, we generate our VQA dataset based on re-grounded reports. 
Here, we employ GPT-4o (\texttt{2024-08-06})~\cite{achiam2023gpt} as a generator due to its remarkable capabilities in understanding and generating long texts. 
\emph{We ensure the quality of the generated dataset by:}
(1) to ensure a diverse range of question content~\cite{liu2021slake}, like ``abnormality'' and ``location'', we identified $7$ distinct categories through discussions with radiologists, as illustrated in Figure~\ref{fig:all_content_distribution}. 
Then, we manually craft questions covering these $7$ types for 30 images, which serve as good demonstrations in the prompt to enhance the generation accuracy and better align with our objectives;
(2) we also design specific rules (like not generating questions that need to be answered by comparing two images) to ensure the generated VQAs are answerable; (3) similar to the re-grounding process, we extract 50 samples to observe the effect of the prompt. Only when the overall performance meets our expectations do we proceed with the final dataset generation. 

For each image-report sample, we instruct the GPT-4o to generate a total of \textbf{11} questions: $3$ open-ended VQAs, $2$ closed-ended VQAs, $3$ single-choice VQAs, and $3$ multi-choice VQAs, culminating in approximately 1.6 million VQA pairs. The reports containing less than three sentences will not be used to generate QAs. The generated location is provided as an anatomical region in text format (\eg, ``left lower lung zone''). To enable the VQA model to identify the specific location on the image, a post-processing step is required to map the region to bounding box coordinates. Figure~\ref{fig:data_construction} illustrates multi-modal explainability, with more examples and the final prompt provided in the Appendix.

\section{Statistics of \shortname{}}

\begin{table}[!t]
    \centering
    \renewcommand{\arraystretch}{1.3}
\resizebox{\linewidth}{!}{
    \begin{tabular}{c|cccc}
    \toprule
    & Open. (T/B) & Closed. (T/B) & Single. (T/B) & Multi. (T/B) \\
    \hline
    Train &  441,471/466,725  & 272,323/277,249  & 441,114/448,810 & 434,067/861,635 \\
    Valid & 3,524/3,704  & 2,145/2,184  & 3,520/3,599  & 3,451/6,955 \\
    Test & 1,144/1,189  & 543/552  & 1,300/1,310  & 973/1,870 \\
    \hline
    Total & 446,139/471,618  & 275,011/279,985  & 445,934/453,719  & 438,491/870,460 \\
    \bottomrule
    \end{tabular}}
    \caption{Distribution statistics of question types (T) and number of bounding boxes (B) across data splits.} 
    \label{tab:vqa_statistics}
\end{table}

\textbf{Dataset Split.} 
\shortname{} is partitioned in accordance with the distribution of MIMIC-CXR. Specifically, we have $149,535$ images with $1,588,975$ QA pairs for training, $1,190$ images with $12,640$ QA pairs for validation, and $300$ images with $3,960$ QA pairs for testing. Detailed statistics, including question type distribution and the number of bounding boxes, are shown in Table~\ref{tab:vqa_statistics}. 


\textbf{Data Quality.} To establish a golden test set, we selected 300 images from the MIMIC-CXR's test set, initially accompanied by 3,291 questions auto-generated by GPT-4o. \emph{Radiologists meticulously reviewed theses questions, correcting around 10 incorrect answers and adjusting 3 inaccurate location annotations. This minimal revision rate demonstrates the high quality of the generated dataset. Additionally, the radiologists contributed approximately 600 new questions, thereby creating a comprehensive golden test set for benchmarking large vision language models.}

\begin{figure}[!t]
  \centering
  \includegraphics[width=1\linewidth]{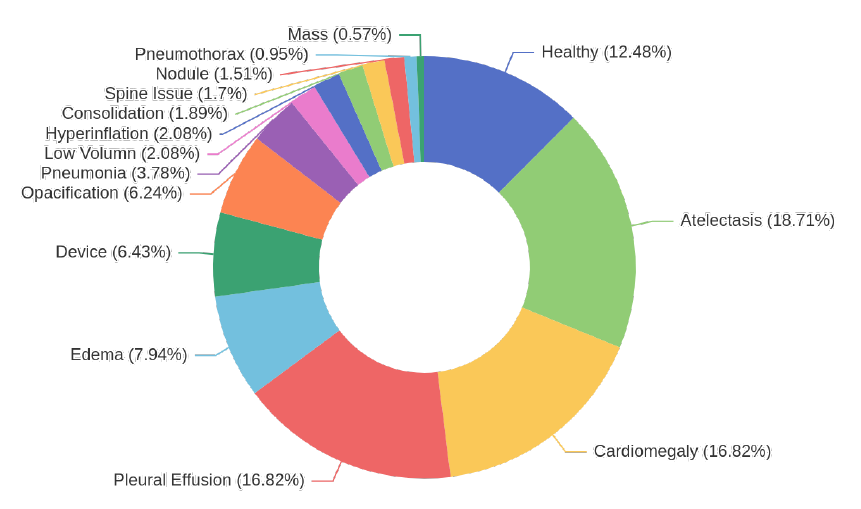}
  \caption{
  The distribution of normality and abnormality contained in images from the test set of our \shortname{}.
  }
  \label{fig:all_test_disease_distribution}
\end{figure}

\textbf{Distribution of Clinical Observations.}
The X-ray image selection for the test set is according to the clinical observations. In Figure~\ref{fig:all_test_disease_distribution}, we plot the distribution of normality and abnormality contained in images. The original distribution of MIMIC-CXR, characterized by a high proportion of healthy samples, introduces a significant bias affecting model performance~\cite{liu2022competence}. 
To mitigate this issue, we preserve only a small proportion of healthy samples (around \textbf{12.48\%}) during manual cleaning. Meanwhile, we ensure that clinically important observations occupy a large proportion, such as ``atelectasis'', ``cardiomegaly'', ``edema'', and ``pleural effusion''~\cite{irvin2019chexpert}. Additionally, common diseases or observations like ``pneumonia'', ``opacification'', and ``pneumothorax'' are also included.

\begin{figure}[!t]
    \centering
    \includegraphics[width=0.85\linewidth]{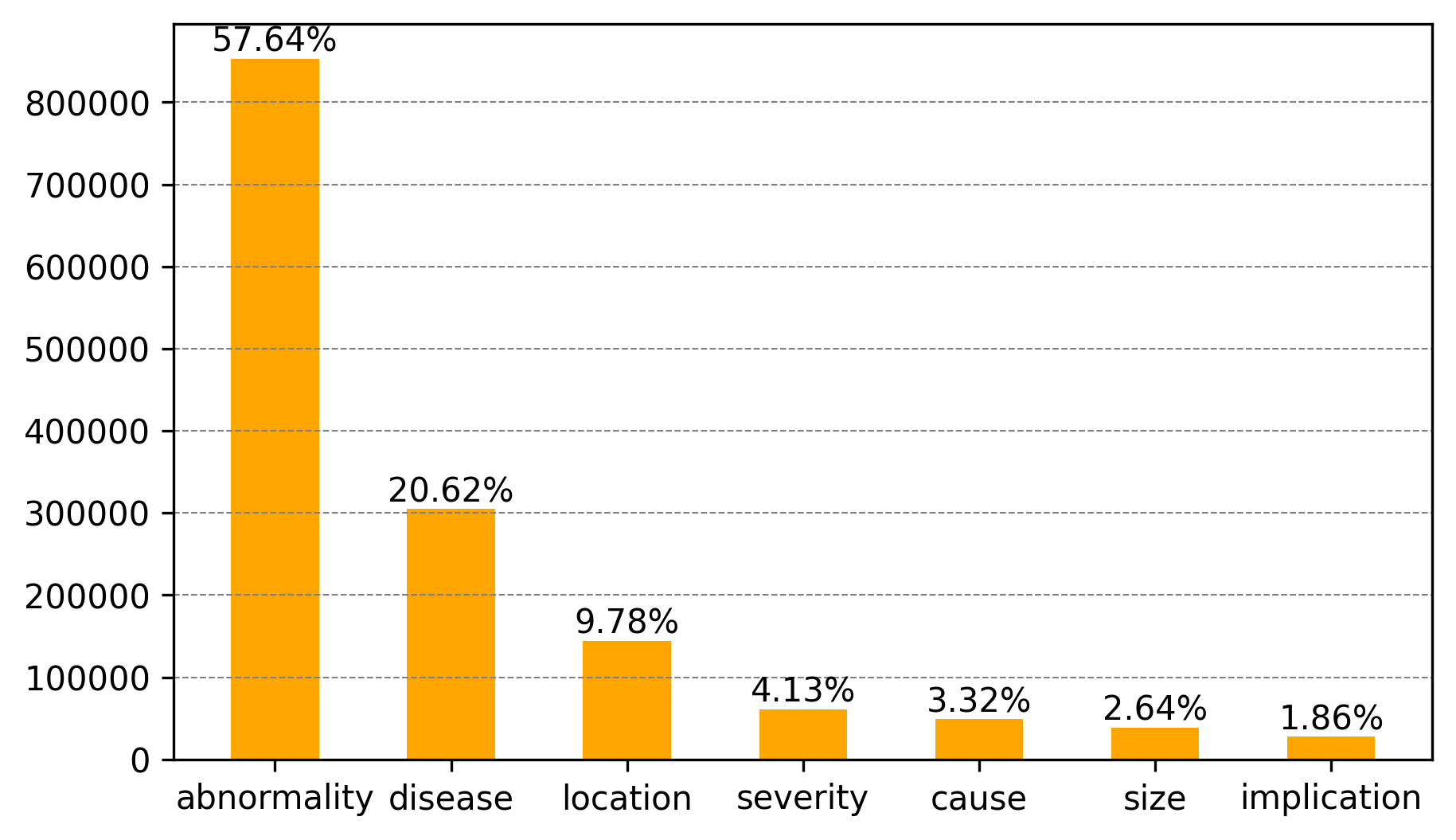}
    \caption{The distribution of question content in our \shortname{}.}
    \label{fig:all_content_distribution}
\end{figure}

\textbf{Distribution of Question Content.}
We show the distribution of the question content categories that GPT-4o itself summarizes according to demonstrations during VQA generation.  Figure~\ref{fig:all_content_distribution} shows the corresponding results where we can find that ``abnormality'', ``disease'', and ``location'' account for over 88\%, while the remaining categories mainly include ``cause'', ``size'', ``severity'', and ``implication'', which highlights the diversity of questions.

More data statistics, such as word cloud and word length distribution, are
provided in the Appendix.

\section{Evaluation }
\subsection{Models}

\paragraph{A Strong Baseline Fine-Tuned on \shortname{}.} To validate the effectiveness of the dataset, especially the auto-generated training set, we propose a question-type-aware instruction tuning to fine-tune LLaVA-Med-v1-7B~\cite{li2024llava} on the training set of \shortname{}, termed as \texttt{LLaVA-Med-\shortname{}}, serving as a strong baseline. Specifically, for each VQA sample from our \shortname{}, we add a type prompt $\bf{X}_{\texttt{Type}}$ after the original system prompt and a question $\bf{X}_{\texttt{Question}}$ with its answer $\bf{X}_{\texttt{Answer}}$, textual reason $\bf{X}_{\texttt{Reason}}$, and corresponding visual location $\bf{X}_{\texttt{Location}}$, constructing a single-turn dialogue as in Table~\ref{tab:prompt_tuning_format}. Generally, $\bf{X}_{\texttt{Type}}$ is ``Input a \{Type\} question, and the assistant will output its answer \{Supplement\} with a detailed reason and corresponding visual location.'' where \{Type\} refers to ``open-ended/closed-ended/single-choice/multi-choice'' and \{Supplement\} is replaced by ``\texttt{none}/(yes or no)/(an option)/(some options)'', respectively. 
Some input samples, as well as the fine-tuning details, are shown in the Appendix.

\begin{table}[!h]\centering
\begin{minipage}{1\columnwidth}\vspace{0mm} \centering
\scalebox{0.80}{
\begin{tcolorbox} 
    \raggedright
    \small
\texttt{<Ori\_System\_Prompt>}  $\bf{X}_{\texttt{Type}}$ \texttt{<STOP>}  \\
\texttt{Human}: \texttt{<image>\textbackslash n}$\bf{X}_{\texttt{Question}}$ $\bf{X}_{\texttt{Choices}}$ (if any) {\texttt{<STOP>}} \\
\texttt{Assistant}: \texttt{<answer>}$\bf{X}_{\texttt{Answer}}$ \texttt{<reason>}$\bf{X}_{\texttt{Reason}}$ \texttt{<location>}$\bf{X}_{\texttt{Location}}$ {\texttt{<STOP>}} \\
 
\end{tcolorbox}
}
\caption{Input format for fine-tuning LLaVA-Med. }
    \label{tab:prompt_tuning_format}
\vspace{-4mm}
\end{minipage}
\end{table}

\begin{table*}[!t]
    \centering
    \small
    \renewcommand{\arraystretch}{1.2}
\resizebox{\linewidth}{!}{
    \begin{tabular}{c|cc|ccc|ccc|ccc|c}
    \toprule
    \hline
    &  \multicolumn{2}{c|}{Open-ended}&\multicolumn{3}{c|}{Closed-ended} & \multicolumn{3}{c|}{Single-choice}& \multicolumn{3}{c|}{Multi-choice} & \\
     Models   &  AR-score\dag & V-score  & A-score & AR-score\dag  & V-score& A-score & AR-score\dag  & V-score & A-score & AR-score\dag  & V-score & Avg.{\dag} \\
     \hline
     Random & - & - & 48.80 & - & - & 25.85 &-& - & 7.50 & - & - &- \\
     \hline
     GPT-4o-mini~\cite{achiam2023gpt} & \textbf{97.68} & \underline{18.05}& 59.30 & 71.14 & \underline{28.64} & \underline{59.00} & \underline{77.47}  & \underline{23.62} & \underline{49.13} & \underline{82.91} & \underline{19.19} & \underline{82.30}\\
     LLaVA-v1~\cite{liu2024visual} & 76.14 & - & 30.76 & 38.02&-&-& 50.47 & - & - &66.52 &- & 57.79\\
     LLaVA-v1.5~\cite{liu2024improved} & 77.62 & - & 58.93 & 57.00 & -& 47.00&  57.05& - & - & 65.17 &- & 64.21\\
     Mini-GPT4-v1~\cite{zhu2023minigpt} & 55.32 & - & 26.33 & 31.09 & - & -& 37.63 & - & - &46.65 & - & 42.67\\
     mPLUG-Owl~\cite{ye2023mplug} & 76.73 & - & 27.26 & 36.70 & - &  32.00 & 46.89 & - & - &67.92 & -& 
 57.06\\
    DeepSeek-VL~\cite{lu2024deepseek} & 79.30 & 11.00 & 57.10 & 59.86 & 8.28 & 51.69  & 62.03 & 8.57  & 17.99 & 70.35 & 12.98 & 67.89\\
    Qwen-VL-Chat~\cite{bai2023qwenvl} & 78.36 & 3.17 & 23.02 & 45.79  & 12.25 & 44.69  & 59.15 & 16.69 & 7.30 & 67.21 &2.26 & 62.63\\
     \hline
     LLaVA-Med-v1~\cite{li2024llava} & 90.34 & - & 62.62 & 69.91& -& - & 61.74 & - & - & 68.14&-&72.53\\
      LLaVA-Med-v1.5~\cite{li2024llava} & 94.43 & - & \underline{71.82} & \underline{76.54} & -& - & 66.04 & - & - & 67.28&-&76.07\\
     MiniGPT-Med~\cite{alkhaldi2024minigpt} &  86.12 & - & 55.24 & 65.25 & - & - & 55.61 & - & - & 64.33 & - & 67.83\\
     XrayGPT~\cite{thawkar2023xraygpt} & 81.17 & - & - & 68.17 & - & - & 48.33 & -  & - & 55.10 & - & 63.19\\
     RadFM~\cite{wu2023towards} & 88.57 & - & 58.01 & 67.91 & - & - & 57.82 & - & - & 62.41 & - & 69.18 \\
     \textbf{LLaVA-Med-\shortname{}} & \underline{97.05} & \textbf{51.47} & \textbf{77.35} & \textbf{80.72} & \textbf{53.20} & \textbf{73.08} & \textbf{81.42} & \textbf{54.57} & \textbf{67.42} & \textbf{84.98} & \textbf{47.99} & \textbf{86.04}\\
    \midrule    
    \bottomrule
    \end{tabular}}
    \caption{Performance of representative LVLMs on our \shortname{} across four different question types. The AR-score combines answer and reasoning to evaluate textual output performance. \dag\ denotes the GPTScore value (\%). The A-score indicates answer or choice accuracy (\%), and the V-score represents mIoU (\%). The best results are bolded, and the second-best are underlined in each column.
    } 
    \label{tab:vqa_result}
\end{table*}

\paragraph{Existing LVLMs.}
Besides fine-tuning a task-oriented model, we perform a zero-shot evaluation on our \shortname{} dataset across the other 12 LVLMs, with 7 in the general domain and the other 5 in the medical domain:
\begin{itemize}
    \item \textbf{In the General Domain:}  we test \texttt{LLaVA-v1}~\cite{liu2024visual}, \texttt{Mini-GPT4-v1}~\cite{zhu2023minigpt}, \texttt{mPLUG-Owl}~\cite{ye2023mplug}, \texttt{LLaVA-v1.5}~\cite{liu2024improved}, \texttt{DeepSeek-VL}~\cite{lu2024deepseek}, \texttt{Qwen-VL-Chat}~\cite{bai2023qwenvl}, and \texttt{GPT-4o-mini (2024-07-18)}. Note that we did not test GPT-4o because its safety protection policy prohibits it from analyzing medical images, especially when asking about the condition of patients.


    \item \textbf{In the Medical Domain:} we evaluate \texttt{LlaVA-Med-v1}~\cite{li2024llava}, \texttt{LLaVA-Med-v1.5}~\cite{li2024llava}, \texttt{MiniGPT-Med}~\cite{alkhaldi2024minigpt}, \texttt{XrayGPT}~\cite{thawkar2023xraygpt}, and \texttt{RadFM}~\cite{wu2023towards}.

\end{itemize}
A detailed introduction can be found in the Appendix.



\subsection{Evaluation Metrics}\label{sec:eval_scores}
In \shortname{}, each question has a corresponding answer, textual reason, and visual location. Ideally, we aim to evaluate all these three aspects with designed metrics as follows:
\begin{itemize}
    \item \textbf{Answer-Reason Score (AR-score)}: In reality, most LVLMs struggle to generate accurate outputs in terms of format. This doesn't necessarily mean these models lack the knowledge to answer the questions but rather simply lack the ability to follow instructions properly. 
    To ensure a fair comparison, we introduce the Answer-Reason score (AR-score) as an evaluation metric for the textual output, where the answer and reason parts from each test sample are merged as a reference (ground truth), and the evaluated LVLM's output serve as a candidate. 
    We use GPTScore~\cite{li2024llava} to calculate the AR-Score from a semantic perspective. Specifically, GPT-4o is leveraged to quantify the correctness by treating the aforementioned reference as a textual response from assistant \#1, while the candidate as the response from assistant \#2. With both responses, the original question, and the X-ray report, GPT-4o assesses the accuracy, relevance, and helpfulness of each assistant's answer and provides an overall score on a scale of 1 to 10, where a higher score indicates better performance. We then calculate the relative score using GPT-4o's reference score for normalization. 
    Besides, we also employ common NLG metrics (\eg, BERTScore~\cite{zhang2019bertscore}, BLEU, ROUGE) to evaluate AR-score.
    
    \item \textbf{Answer Score (A-score)}: For responses where the model can output specific answers (such as yes/no for closed-ended questions or options for single/multiple choice questions), we calculate the accuracy by comparing with the ground truth. It is worth noting that although some models cannot directly output the answer, we still attempt to match it from their responses.

    \item \textbf{Visual Score (V-score)}: For models capable of visual grounding (\ie, outputting visual locations), we calculated mean intersection over union (mIoU) as a measurement. For a VQA case, considering there might be multiple corresponding locations (commonly seen in multi-choice questions), we use the Hungarian algorithm~\cite{kuhn1955hungarian} to match the predicted bounding boxes with the actual ones.
    \end{itemize}

\begin{table*}[t]
    \centering
    \renewcommand{\arraystretch}{1.2}
\resizebox{\linewidth}{!}{
    \begin{tabular}{c|ccc|ccc|ccc|ccc}
    \toprule
    \hline
    &  \multicolumn{3}{c|}{Open-ended}&\multicolumn{3}{c|}{Closed-ended} & \multicolumn{3}{c|}{Single-choice}& \multicolumn{3}{c}{Multi-choice} \\
     Models   &  BERTScore & ROUGE-L  & BLEU-1 & BERTScore & ROUGE-L  & BLEU-1& BERTScore & ROUGE-L  & BLEU-1 & BERTScore & ROUGE-L  & BLEU-1\\
     \hline
     GPT-4o-mini~\cite{achiam2023gpt} &\underline{30.43} & \underline{22.67} & \underline{18.25} &40.02 & 25.63 & 19.10 & \underline{48.34} & \underline{39.17} & \underline{30.82} &  \underline{46.58} & \underline{39.20} & \underline{28.65}\\
     LLaVA-v1~\cite{liu2024visual} & 20.09 & 15.22 & 11.57 & 22.42 & 13.10 & 8.01 & 20.25 & 14.97 & 10.61 & 19.69 & 17.35 & 11.15\\
     LLaVA-v1.5~\cite{liu2024improved} & 21.49 & 16.11 & 12.20 & 32.59 & 15.37 & 6.69 & 17.42 & 17.53 & 1.49 & 23.74 & 21.20 & 8.95\\
     Mini-GPT4-v1~\cite{zhu2023minigpt} & 15.03 & 14.66 & 11.46 & 13.83 & 9.65 & 6.31 & 6.50 & 6.79 & 4.60 &  5.31 & 5.79 & 3.22\\
     mPLUG-Owl~\cite{ye2023mplug} & 22.52 & 17.03 & 13.22 & 32.23 & 20.20 & 13.92 & 39.64 & 33.69 & 30.32 & 26.09 & 24.97 & 16.68 \\
    DeepSeek-VL~\cite{lu2024deepseek} & 24.06 & 18.62 & 15.94 &26.12 & 23.27 &13.83  & 26.16 & 30.46 & 18.79 & 22.10 & 27.32 & 20.74  \\
    Qwen-VL-Chat~\cite{bai2023qwenvl} & 23.31 & 18.48  &  14.63 & 33.18 & 22.43 & 17.19 & 25.47 & 22.25 & 6.95 & 22.03& 22.88 & 12.26 \\
     \hline
     LLaVA-Med-v1~\cite{li2024llava} & 25.14 & 19.63 & 15.93 & 38.04 & 29.08 & 19.74 & 34.89 & 30.10 & 25.84 & 28.63 & 26.51 & 20.99 \\
      LLaVA-Med-v1.5~\cite{li2024llava} & 26.42 & 21.38 & 17.28 & \underline{44.48} & \underline{36.73} & \underline{26.35} & 36.62 & 30.32 & 25.44 & 28.11 & 24.49 & 16.53 \\
     MiniGPT-Med~\cite{alkhaldi2024minigpt} & 23.47 & 19.20 & 16.03 & 34.31 & 29.47 & 19.13 & 30.11 & 28.51 & 22.13 & 26.51 & 24.42 & 15.98 \\
     XrayGPT~\cite{thawkar2023xraygpt} &  22.57 & 18.30 & 15.73 & 21.35 & 14.55 & 10.17 & 16.31 &  12.17 & 9.23 & 12.15 & 10.30 & 6.22\\
     RadFM~\cite{wu2023towards} & 24.96 & 20.71 & 17.73 & 37.43 & 27.95 & 20.56 & 32.30 & 27.02 & 24.39 & 25.81 &  20.02 & 13.80\\
     \textbf{LLaVA-Med-\shortname{}} & \textbf{42.69} & \textbf{32.75} & \textbf{25.28} & \textbf{54.44} & \textbf{38.39} & \textbf{33.99} & \textbf{56.35} & \textbf{53.23} & \textbf{47.31} &\textbf{54.95}& \textbf{50.85}& \textbf{43.99}  \\
    \midrule    
    \bottomrule
    \end{tabular}}
    \caption{
    Performance of representative LVLMs evaluated using various natural language generation metrics for AR-score, including BERTScore, ROUGE-L, and BLEU-1. The best results are bolded, and the second-best are underlined in each column.
    } 
    \label{tab:vqa_nlg_result}
\end{table*}

\subsection{Results and Analysis}

\paragraph{Overall Performance.} 
The comprehensive results are shown in Table~\ref{tab:vqa_result}. 
The first 7 rows indicate the performance of general LVLMs, while the last 6 rows present the results of medical ones and our fine-tuned version of LLaVA-Med-v1 (termed as LLaVA-Med-\shortname{}). It can be found that:
\begin{itemize}
    \item \textbf{Most exisiting LVLMs exhibit weak performance when tested on \shortname{}.} The only exception is GPT-4o-mini, which achieves an AR-score above 80 on average across all tasks. When considering specific question types, LLaVA-Med (both versions 1 and 1.5) stands out for its strong performance on open-ended questions, scoring above 90 on the AR-score. However, all models show poor results on the other three categories of tasks.

    \item \textbf{When faced with choice-based questions, most models, particularly those in the medical domain, struggle to provide definitive answers, despite their ability to analyze each option.} This difficulty accounts for why many models have an associated AR-score but lack an A-score, highlighting the importance of introducing these types of questions.


    \item \textbf{Powerful LVLMs, such as GPT-4o-mini, often rely on shortcut reasoning rather than real multimodal reasoning}. Although these models can sometimes answer questions to a certain extent (as indicated by the AR-scores), they often fail to accurately achieve visual grounding. This suggests that these models tend to address Med-VQA tasks using shortcut knowledge, such as retrieving information from their pre-training memory, instead of engaging in real multimodal reasoning~\cite{chen2024detecting}. However, real multimodal reasoning is essential for the explainability of Med-VQA systems.

    \item \textbf{Through simple question-type-aware instruction tuning, the proposed baseline model achieves a significant performance improvement}, with an approximate \textbf{13.5\%} increase in average AR-score compared to LLaVA-Med-v1. Notably, it surpasses GPT-4o-mini on most metrics, demonstrating the reliability of the training set. However, 
    a substantial gap remains for practical application, highlighting the challenges associated with \shortname{}.

\end{itemize}

\textbf{Limitation.} Note that the proposed baseline model is inherently task-specific, which may result in reduced accuracy on other tasks or a diminished capacity for conversational engagement. The true potential of our \shortname{} lies in its integration into multi-task training frameworks, such as the second training stage of LLaVA-Med. The baseline model primarily serves to demonstrate the dataset's effectiveness while also providing a robust benchmark.


\paragraph{More Metrics.} As mentioned in Section~\ref{sec:eval_scores}, we also calculate NLG metrics to measure AR-score. Detailed results are shown in Table~\ref{tab:vqa_nlg_result}. Overall, the NLG metrics generally share the same trend as GPTScore (AR-score in Table~\ref{tab:vqa_result}), but there are some minor differences. 
(1) \textbf{
High NLG scores do not always correlate with strong model performance}, as seen with mPLUG-Owl compared to LLaVA-v1.5. Essentially, LLaVA-1.5 demonstrates higher performance, such as achieving an answer accuracy rate (A-score) in single-choice tasks that is \textbf{15\%} higher than that of mPLUG-Owl. However, since LLaVA-v1.5’s output mostly consists of the answer without reason, the shorter output results in a lower NLG score, with its BLEU-1 approximately \textbf{28.8\%} lower than mPLUG-Owl;
(2) \textbf{
Fine-tuning on \shortname{} results in more pronounced improvements in NLG metrics compared to GPTScore}. 
For example, the fine-tuned baseline model shows only a 3.7\% average improvement over GPT-4o-mini on GPTScore, but achieves a 12.1\% improvement on average NLG metrics. This significant enhancement better reflects the model's learning from the dataset. We suggest using GPTScore for models not fine-tuned on \shortname{} due to its focus on semantic understanding, while NLG metrics are preferred for fine-tuned models as they better capture alignment with the dataset.


\begin{table*}[!t]
  \begin{minipage}{0.99\textwidth}
\centering  
\vspace{-4mm}
\scalebox{0.70}{
\small
\begin{tabular}{l |p{6.5cm}| p{6.5cm} | p{6.5cm} }
\toprule
 \multicolumn{3}{l}{\bf Challenging examples from \shortname{}:}  \\
\midrule
& \textbf{CASE I} & \textbf{CASE II} & \textbf{CASE III}\\
&  \includegraphics[height=4.5cm]{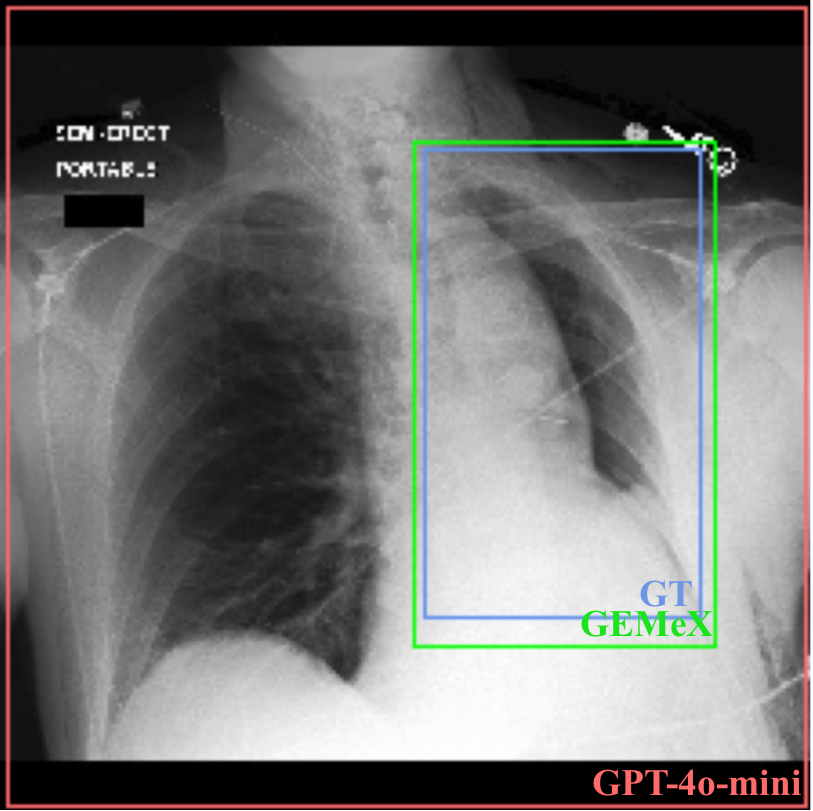} &  \includegraphics[height=4.5cm]{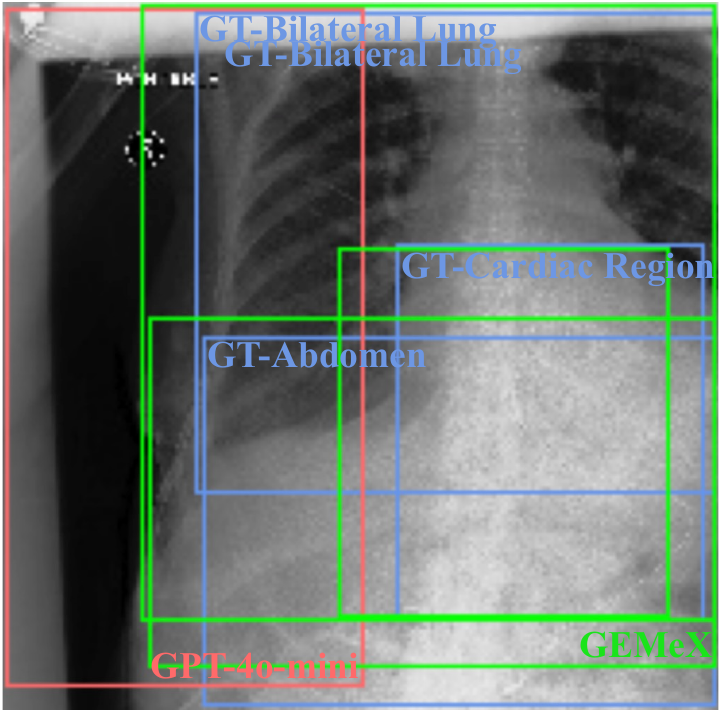} &
\includegraphics[height=4.5cm]{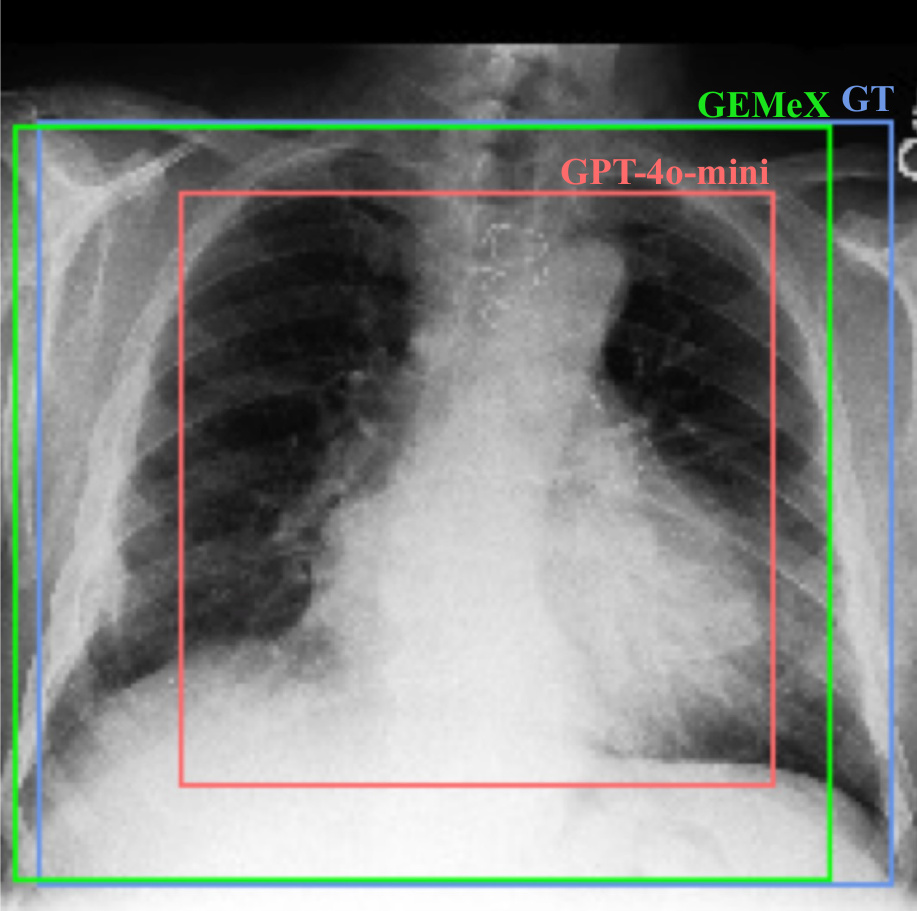}\\
& (Bbox predicted by each model) & (Bbox predicted by each model) &(Bbox predicted by each model)\\
 \midrule
\emph{Question} & What might be the cause of the overexpansion of the right lung?  & What regions on the X-ray indicate abnormal findings? Choices: [A: Bilateral lung, B: Cardiac region, C: Abdomen, D: Spine] & What is the overall condition of the lungs based on the CXR? Choices: [A: Clear, B: Consolidated, C: Effused, D: Thickened]\\
\emph{Ground Truth} & \textless Answer\textgreater: \textcolor{gray}{Compensatory hyperinflation} due to left lung removal. \textless Reason\textgreater: After a pneumonectomy, the remaining lung may overexpand to compensate for the \textcolor{orange}{loss of the left lung}.  & \textless Answer\textgreater: [A, B, C]. \textless Reason\textgreater: \textcolor{red}{The bilateral lung shows pleural effusion}, \textcolor{gray}{the cardiac region shows enlargement}, and \textcolor{orange}{the abdomen shows the nasogastric tube}. The spine is normal. & \textless Answer\textgreater: A.  \textless Reason\textgreater: The CXR indicates that \textcolor{orange}{the lungs are clear of consolidation or effusion, although right pleural-based thickening at the base}. \\

\\ \midrule
\texttt{GPT-4o-mini} &  \textless Answer\textgreater: The overexpansion of the right lung may be caused by conditions such as a right-sided pneumothorax or chronic obstructive pulmonary disease (COPD).   \textless Reason\textgreater: Overexpansion can occur when air enters the pleural space (as in pneumothorax) or when there is an obstruction that prevents air from leaving the lung, causing it to become hyperinflated. In COPD, the lung tissue loses elasticity, leading to overinflation. \textless GPTScore\textgreater: 44.44 & \textless Answer\textgreater: [A: Bilateral lung]. \textless Reason\textgreater: \textcolor{red}{The bilateral lung region shows potential abnormalities such as opacities or other irregularities that may indicate pathology (\XSolidBrush)}. \textless GPTScore\textgreater: 37.50 & \textless Answer\textgreater: B. \textless Reason\textgreater: \textcolor{orange}{The presence of increased opacity in the lung fields suggests consolidation, indicating that the lungs are not clear and may have fluid or solid material in the alveoli (\XSolidBrush)}. \textless GPTScore\textgreater: 33.33.
\\ \midrule
\texttt{LLaVA-Med-\shortname{}} & \textless Answer\textgreater: Possible chronic lung disease or \textcolor{gray}{compensatory mechanism (\Checkmark)}. \textless Reason\textgreater: Overexpansion of the lung can be due to chronic lung diseases or compensatory mechanisms, especially \textcolor{orange}{after a left lobectomy (\Checkmark)}. \textless GPTScore\textgreater: 66.67 & \textless Answer\textgreater: [A, B, C]. \textless Reason\textgreater: \textcolor{red}{The bilateral lung shows interstitial opacities (\XSolidBrush)}, \textcolor{gray}{the cardiac region shows enlargement (\Checkmark)} , and \textcolor{orange}{the abdomen shows the feeding tube location (\Checkmark)}.  \textless GPTScore\textgreater: 87.50 & \textless Answer\textgreater: A. \textless Reason\textgreater: The CXR indicates that \textcolor{orange}{the lungs are otherwise clear except for the noted pleural thickening (\Checkmark)}. \textless GPTScore\textgreater: 88.89\\
\\ \bottomrule
\end{tabular}
}
\vspace{1mm}
\captionof{table}{
Challenging questions from \shortname{} answered by GPT-4o-mini and our LLaVA-Med-\shortname{}, \textbf{highlighting the limitations of LVLMs like GPT-4o-mini in visual reasoning, which can be enhanced by integrating knowledge from \shortname{}}. Colors in outputs highlight pathological descriptions that are similar to the ground truth, though they may be either correct (\Checkmark) or incorrect (\XSolidBrush).}
\label{tab:cases_slake_cxr}  
  \end{minipage}
\end{table*}

\begin{table}[!t]
    \centering
\resizebox{0.95\linewidth}{!}{
    \begin{tabular}{c|c|cc}
    \toprule
    \hline
    Models & Open-ended &  \multicolumn{2}{c}{Closed-ended}\\
    & AR-Score & A-Score & AR-Score \\
    \hline
    LLaVA-Med-v1 &  73.31   & 56.17  &  62.35\\
    \textbf{LLaVA-Med-\shortname{}} &  \textbf{82.78}   &  \textbf{69.79} & \textbf{75.06}  \\
    \hline
    \bottomrule
    \end{tabular}
    }
\caption{Performance of LVLMs on SLAKE-CXR.} 
    \label{tab:transferability}
\end{table}
\paragraph{Transferability.} To further validate the effectiveness of \shortname{}, we assess it from a transfer learning perspective by treating \shortname{} as a pre-training dataset and subsequently testing our LLaVA-Med-\shortname{} model on other X-ray VQA datasets. 
Considering potential overlap in data sources, we utilize the CXR portion of the SLAKE test set~\cite{liu2021slake}, which includes 505 open-ended questions and 235 closed-ended questions. The zero-shot performance of LLaVA-Med-\shortname{} and LLaVA-Med-v1 on this dataset is reported in Table~\ref{tab:transferability}.
The results indicate that our LLaVA-Med-\shortname{} significantly outperforms LLaVA-Med-v1 in both tasks, demonstrating the effective X-ray knowledge acquired from \shortname{}.

\paragraph{Case Study.} In Table~\ref{tab:cases_slake_cxr}, we show some questions with outputs from both GPT-4o-mini and our fine-tuned LLaVA-Med-\shortname{} for qualitative comparison.  In CASE I, although GPT-4o-mini can generate a very detailed answer, it provides answers without reasoning on the visual content, resulting in a significant difference from the ground truth. In contrast, the LLaVA-Med-\shortname{} offers relatively accurate visual clues and is able to provide partially correct answers (``the compensatory mechanism''), although there is a false mention of ``possible chronic lung disease'' when considering the patient's condition. In CASE II, although GPT-4o-mini can analyze images, its limited capabilities result in selecting only one option and providing a vague reason. In contrast, the LLaVA-Med-\shortname{} outputs the correct options but gives an incorrect reason for one option (\ie, answer ``A''). In CASE III,  GPT-4o-mini cannot both visually reason and output answers correctly, while the fine-tuned model can give better outputs from these two aspects. From these examples, we can conclude that some LVLMs still lack sufficient understanding of medical images. Meanwhile, while the proposed simple fine-tuning method improves performance, it is still far from fully accurate, leaving much room for further exploration. We have provided more case studies in the Appendix.

    

\section{Conclusion}
In this paper, we introduce a benchmark, \shortname{}, designed to advance the field of medical VQA with two primary advantages: multimodal explainability and diverse question types. 
\shortname{} not only provides more accessible medical explanations for patients and junior doctors but also serves as a valuable training resource for developing next-generation medical LVLMs with enhanced instruction-following capabilities. 
We demonstrate the effectiveness and difficulty of the dataset through comprehensive testing of representative LVLMs as well as task-specific fine-tuning, hoping that \shortname{} can promote medical VQA development and improve AI-assisted medical care.
{
    \small
    \bibliographystyle{ieeenat_fullname}
    \bibliography{main}

\begin{thebibliography}{56}
\providecommand{\natexlab}[1]{#1}
\providecommand{\url}[1]{\texttt{#1}}
\expandafter\ifx\csname urlstyle\endcsname\relax
  \providecommand{\doi}[1]{doi: #1}\else
  \providecommand{\doi}{doi: \begingroup \urlstyle{rm}\Url}\fi

\bibitem[Achiam et~al.(2023)Achiam, Adler, Agarwal, Ahmad, Akkaya, Aleman, Almeida, Altenschmidt, Altman, Anadkat, et~al.]{achiam2023gpt}
Josh Achiam, Steven Adler, Sandhini Agarwal, Lama Ahmad, Ilge Akkaya, Florencia~Leoni Aleman, Diogo Almeida, Janko Altenschmidt, Sam Altman, Shyamal Anadkat, et~al.
\newblock Gpt-4 technical report.
\newblock \emph{arXiv preprint arXiv:2303.08774}, 2023.

\bibitem[Alayrac et~al.(2022)Alayrac, Donahue, Luc, Miech, Barr, Hasson, Lenc, Mensch, Millican, Reynolds, et~al.]{alayrac2022flamingo}
Jean-Baptiste Alayrac, Jeff Donahue, Pauline Luc, Antoine Miech, Iain Barr, Yana Hasson, Karel Lenc, Arthur Mensch, Katherine Millican, Malcolm Reynolds, et~al.
\newblock Flamingo: a visual language model for few-shot learning.
\newblock \emph{Advances in neural information processing systems}, 35:\penalty0 23716--23736, 2022.

\bibitem[Alkhaldi et~al.(2024)Alkhaldi, Alnajim, Alabdullatef, Alyahya, Chen, Zhu, Alsinan, and Elhoseiny]{alkhaldi2024minigpt}
Asma Alkhaldi, Raneem Alnajim, Layan Alabdullatef, Rawan Alyahya, Jun Chen, Deyao Zhu, Ahmed Alsinan, and Mohamed Elhoseiny.
\newblock Minigpt-med: Large language model as a general interface for radiology diagnosis.
\newblock \emph{arXiv preprint arXiv:2407.04106}, 2024.

\bibitem[Bae et~al.(2024)Bae, Kyung, Ryu, Cho, Lee, Kweon, Oh, Ji, Chang, Kim, et~al.]{bae2024ehrxqa}
Seongsu Bae, Daeun Kyung, Jaehee Ryu, Eunbyeol Cho, Gyubok Lee, Sunjun Kweon, Jungwoo Oh, Lei Ji, Eric Chang, Tackeun Kim, et~al.
\newblock Ehrxqa: A multi-modal question answering dataset for electronic health records with chest x-ray images.
\newblock \emph{Advances in Neural Information Processing Systems}, 36, 2024.

\bibitem[Bai et~al.(2023{\natexlab{a}})Bai, Bai, Yang, Wang, Tan, Wang, Lin, Zhou, and Zhou]{bai2023qwen}
Jinze Bai, Shuai Bai, Shusheng Yang, Shijie Wang, Sinan Tan, Peng Wang, Junyang Lin, Chang Zhou, and Jingren Zhou.
\newblock Qwen-vl: A versatile vision-language model for understanding, localization, text reading, and beyond.
\newblock \emph{arXiv preprint arXiv:2308.12966}, 1\penalty0 (2):\penalty0 3, 2023{\natexlab{a}}.

\bibitem[Bai et~al.(2023{\natexlab{b}})Bai, Bai, Yang, Wang, Tan, Wang, Lin, Zhou, and Zhou]{bai2023qwenvl}
Jinze Bai, Shuai Bai, Shusheng Yang, Shijie Wang, Sinan Tan, Peng Wang, Junyang Lin, Chang Zhou, and Jingren Zhou.
\newblock Qwen-vl: A versatile vision-language model for understanding, localization, text reading, and beyond, 2023{\natexlab{b}}.

\bibitem[Ben~Abacha et~al.(2021)Ben~Abacha, Sarrouti, Demner-Fushman, Hasan, and M{\"u}ller]{ben2021overview}
Asma Ben~Abacha, Mourad Sarrouti, Dina Demner-Fushman, Sadid~A Hasan, and Henning M{\"u}ller.
\newblock Overview of the vqa-med task at imageclef 2021: Visual question answering and generation in the medical domain.
\newblock In \emph{Proceedings of the CLEF 2021 Conference and Labs of the Evaluation Forum-working notes}. 21-24 September 2021, 2021.

\bibitem[Chen et~al.(2024{\natexlab{a}})Chen, Yang, Wu, Jiang, Hou, Li, Wang, Xiao, Li, and Zhang]{chen2024detecting}
Jiawei Chen, Dingkang Yang, Tong Wu, Yue Jiang, Xiaolu Hou, Mingcheng Li, Shunli Wang, Dongling Xiao, Ke Li, and Lihua Zhang.
\newblock Detecting and evaluating medical hallucinations in large vision language models.
\newblock \emph{arXiv preprint arXiv:2406.10185}, 2024{\natexlab{a}}.

\bibitem[Chen et~al.(2024{\natexlab{b}})Chen, Zhao, Zhang, Xu, Gao, Xu, Wu, Li, Shi, and He]{chen2024eyegpt}
Xiaolan Chen, Ziwei Zhao, Weiyi Zhang, Pusheng Xu, Le Gao, Mingpu Xu, Yue Wu, Yinwen Li, Danli Shi, and Mingguang He.
\newblock Eyegpt: Ophthalmic assistant with large language models.
\newblock \emph{arXiv preprint arXiv:2403.00840}, 2024{\natexlab{b}}.

\bibitem[Chen et~al.(2022)Chen, Du, Hu, Liu, Li, Wan, and Chang]{chen2022multi}
Zhihong Chen, Yuhao Du, Jinpeng Hu, Yang Liu, Guanbin Li, Xiang Wan, and Tsung-Hui Chang.
\newblock Multi-modal masked autoencoders for medical vision-and-language pre-training.
\newblock In \emph{International Conference on Medical Image Computing and Computer-Assisted Intervention}, pages 679--689. Springer, 2022.

\bibitem[Chen et~al.(2023)Chen, Diao, Wang, Li, and Wan]{chen2023towards}
Zhihong Chen, Shizhe Diao, Benyou Wang, Guanbin Li, and Xiang Wan.
\newblock Towards unifying medical vision-and-language pre-training via soft prompts.
\newblock In \emph{Proceedings of the IEEE/CVF International Conference on Computer Vision}, pages 23403--23413, 2023.

\bibitem[Chiang et~al.(2023)Chiang, Li, Lin, Sheng, Wu, Zhang, Zheng, Zhuang, Zhuang, Gonzalez, et~al.]{chiang2023vicuna}
Wei-Lin Chiang, Zhuohan Li, Zi Lin, Ying Sheng, Zhanghao Wu, Hao Zhang, Lianmin Zheng, Siyuan Zhuang, Yonghao Zhuang, Joseph~E Gonzalez, et~al.
\newblock Vicuna: An open-source chatbot impressing gpt-4 with 90\%* chatgpt quality.
\newblock \emph{See https://vicuna. lmsys. org (accessed 14 April 2023)}, 2\penalty0 (3):\penalty0 6, 2023.

\bibitem[Dai et~al.(2023)Dai, Li, Li, Tiong, Zhao, Wang, Li, Fung, and Hoi]{dai2023instructblip}
Wenliang Dai, Junnan Li, Dongxu Li, Anthony Meng~Huat Tiong, Junqi Zhao, Weisheng Wang, Boyang Li, Pascale Fung, and Steven Hoi.
\newblock Instructblip: Towards general-purpose vision-language models with instruction tuning, 2023.

\bibitem[Gong et~al.(2021)Gong, Chen, Liu, Yu, and Li]{gong2021cross}
Haifan Gong, Guanqi Chen, Sishuo Liu, Yizhou Yu, and Guanbin Li.
\newblock Cross-modal self-attention with multi-task pre-training for medical visual question answering.
\newblock In \emph{Proceedings of the 2021 international conference on multimedia retrieval}, pages 456--460, 2021.

\bibitem[Gu et~al.(2024)Gu, Zhu, Zhu, Chen, Tang, and Wang]{gu2024anomalygpt}
Zhaopeng Gu, Bingke Zhu, Guibo Zhu, Yingying Chen, Ming Tang, and Jinqiao Wang.
\newblock Anomalygpt: Detecting industrial anomalies using large vision-language models.
\newblock In \emph{Proceedings of the AAAI Conference on Artificial Intelligence}, pages 1932--1940, 2024.

\bibitem[He et~al.(2020)He, Zhang, Mou, Xing, and Xie]{he2020pathvqa}
Xuehai He, Yichen Zhang, Luntian Mou, Eric Xing, and Pengtao Xie.
\newblock Pathvqa: 30000+ questions for medical visual question answering.
\newblock \emph{arXiv preprint arXiv:2003.10286}, 2020.

\bibitem[Hu et~al.(2023)Hu, Gu, An, Zhang, Liu, Kobayashi, Harada, Summers, and Zhu]{hu2023expert}
Xinyue Hu, Lin Gu, Qiyuan An, Mengliang Zhang, Liangchen Liu, Kazuma Kobayashi, Tatsuya Harada, Ronald~M Summers, and Yingying Zhu.
\newblock Expert knowledge-aware image difference graph representation learning for difference-aware medical visual question answering.
\newblock In \emph{Proceedings of the 29th ACM SIGKDD Conference on Knowledge Discovery and Data Mining}, pages 4156--4165, 2023.

\bibitem[Hu et~al.(2024)Hu, Li, Lu, Shao, He, Qiao, and Luo]{hu2024omnimedvqa}
Yutao Hu, Tianbin Li, Quanfeng Lu, Wenqi Shao, Junjun He, Yu Qiao, and Ping Luo.
\newblock Omnimedvqa: A new large-scale comprehensive evaluation benchmark for medical lvlm.
\newblock In \emph{Proceedings of the IEEE/CVF Conference on Computer Vision and Pattern Recognition}, pages 22170--22183, 2024.

\bibitem[Irvin et~al.(2019)Irvin, Rajpurkar, Ko, Yu, Ciurea-Ilcus, Chute, Marklund, Haghgoo, Ball, Shpanskaya, et~al.]{irvin2019chexpert}
Jeremy Irvin, Pranav Rajpurkar, Michael Ko, Yifan Yu, Silviana Ciurea-Ilcus, Chris Chute, Henrik Marklund, Behzad Haghgoo, Robyn Ball, Katie Shpanskaya, et~al.
\newblock Chexpert: A large chest radiograph dataset with uncertainty labels and expert comparison.
\newblock In \emph{Proceedings of the AAAI conference on artificial intelligence}, pages 590--597, 2019.

\bibitem[Jiang et~al.(2023)Jiang, Sablayrolles, Mensch, Bamford, Chaplot, Casas, Bressand, Lengyel, Lample, Saulnier, et~al.]{jiang2023mistral}
Albert~Q Jiang, Alexandre Sablayrolles, Arthur Mensch, Chris Bamford, Devendra~Singh Chaplot, Diego de~las Casas, Florian Bressand, Gianna Lengyel, Guillaume Lample, Lucile Saulnier, et~al.
\newblock Mistral 7b.
\newblock \emph{arXiv preprint arXiv:2310.06825}, 2023.

\bibitem[Johnson et~al.(2019)Johnson, Pollard, Berkowitz, Greenbaum, Lungren, Deng, Mark, and Horng]{johnson2019mimic}
Alistair~EW Johnson, Tom~J Pollard, Seth~J Berkowitz, Nathaniel~R Greenbaum, Matthew~P Lungren, Chih-ying Deng, Roger~G Mark, and Steven Horng.
\newblock Mimic-cxr, a de-identified publicly available database of chest radiographs with free-text reports.
\newblock \emph{Scientific data}, 6\penalty0 (1):\penalty0 317, 2019.

\bibitem[Khare et~al.(2021)Khare, Bagal, Mathew, Devi, Priyakumar, and Jawahar]{khare2021mmbert}
Yash Khare, Viraj Bagal, Minesh Mathew, Adithi Devi, U~Deva Priyakumar, and CV Jawahar.
\newblock Mmbert: Multimodal bert pretraining for improved medical vqa.
\newblock In \emph{2021 IEEE 18th International Symposium on Biomedical Imaging (ISBI)}, pages 1033--1036. IEEE, 2021.

\bibitem[Kuhn(1955)]{kuhn1955hungarian}
Harold~W Kuhn.
\newblock The hungarian method for the assignment problem.
\newblock \emph{Naval research logistics quarterly}, 2\penalty0 (1-2):\penalty0 83--97, 1955.

\bibitem[Lau et~al.(2018)Lau, Gayen, Ben~Abacha, and Demner-Fushman]{lau2018dataset}
Jason~J Lau, Soumya Gayen, Asma Ben~Abacha, and Dina Demner-Fushman.
\newblock A dataset of clinically generated visual questions and answers about radiology images.
\newblock \emph{Scientific data}, 5\penalty0 (1):\penalty0 1--10, 2018.

\bibitem[Lee et~al.(2023)Lee, Kim, Chang, and Ye]{lee2023llm}
Suhyeon Lee, Won~Jun Kim, Jinho Chang, and Jong~Chul Ye.
\newblock Llm-cxr: Instruction-finetuned llm for cxr image understanding and generation.
\newblock \emph{arXiv preprint arXiv:2305.11490}, 2023.

\bibitem[Li et~al.(2024)Li, Wong, Zhang, Usuyama, Liu, Yang, Naumann, Poon, and Gao]{li2024llava}
Chunyuan Li, Cliff Wong, Sheng Zhang, Naoto Usuyama, Haotian Liu, Jianwei Yang, Tristan Naumann, Hoifung Poon, and Jianfeng Gao.
\newblock Llava-med: Training a large language-and-vision assistant for biomedicine in one day.
\newblock \emph{Advances in Neural Information Processing Systems}, 36, 2024.

\bibitem[Li et~al.(2018)Li, Tao, Joty, Cai, and Luo]{li2018vqa}
Qing Li, Qingyi Tao, Shafiq Joty, Jianfei Cai, and Jiebo Luo.
\newblock Vqa-e: Explaining, elaborating, and enhancing your answers for visual questions.
\newblock In \emph{Proceedings of the European Conference on Computer Vision (ECCV)}, pages 552--567, 2018.

\bibitem[Lin et~al.(2023{\natexlab{a}})Lin, Zhao, Zhang, Wu, Zhang, Wang, and Xie]{lin2023pmc}
Weixiong Lin, Ziheng Zhao, Xiaoman Zhang, Chaoyi Wu, Ya Zhang, Yanfeng Wang, and Weidi Xie.
\newblock Pmc-clip: Contrastive language-image pre-training using biomedical documents.
\newblock In \emph{International Conference on Medical Image Computing and Computer-Assisted Intervention}, pages 525--536. Springer, 2023{\natexlab{a}}.

\bibitem[Lin et~al.(2023{\natexlab{b}})Lin, Zhang, Tao, Shi, Haffari, Wu, He, and Ge]{lin2023medical}
Zhihong Lin, Donghao Zhang, Qingyi Tao, Danli Shi, Gholamreza Haffari, Qi Wu, Mingguang He, and Zongyuan Ge.
\newblock Medical visual question answering: A survey.
\newblock \emph{Artificial Intelligence in Medicine}, 143:\penalty0 102611, 2023{\natexlab{b}}.

\bibitem[Liu et~al.(2021)Liu, Zhan, Xu, Ma, Yang, and Wu]{liu2021slake}
Bo Liu, Li-Ming Zhan, Li Xu, Lin Ma, Yan Yang, and Xiao-Ming Wu.
\newblock Slake: A semantically-labeled knowledge-enhanced dataset for medical visual question answering.
\newblock In \emph{2021 IEEE 18th International Symposium on Biomedical Imaging (ISBI)}, pages 1650--1654. IEEE, 2021.

\bibitem[Liu et~al.(2022{\natexlab{a}})Liu, Zhan, Xu, and Wu]{liu2022medical}
Bo Liu, Li-Ming Zhan, Li Xu, and Xiao-Ming Wu.
\newblock Medical visual question answering via conditional reasoning and contrastive learning.
\newblock \emph{IEEE transactions on medical imaging}, 42\penalty0 (5):\penalty0 1532--1545, 2022{\natexlab{a}}.

\bibitem[Liu et~al.(2022{\natexlab{b}})Liu, Ge, Zou, and Wu]{liu2022competence}
Fenglin Liu, Shen Ge, Yuexian Zou, and Xian Wu.
\newblock Competence-based multimodal curriculum learning for medical report generation.
\newblock \emph{arXiv preprint arXiv:2206.14579}, 2022{\natexlab{b}}.

\bibitem[Liu et~al.(2024{\natexlab{a}})Liu, Li, Li, and Lee]{liu2024improved}
Haotian Liu, Chunyuan Li, Yuheng Li, and Yong~Jae Lee.
\newblock Improved baselines with visual instruction tuning.
\newblock In \emph{Proceedings of the IEEE/CVF Conference on Computer Vision and Pattern Recognition}, pages 26296--26306, 2024{\natexlab{a}}.

\bibitem[Liu et~al.(2024{\natexlab{b}})Liu, Li, Wu, and Lee]{liu2024visual}
Haotian Liu, Chunyuan Li, Qingyang Wu, and Yong~Jae Lee.
\newblock Visual instruction tuning.
\newblock \emph{Advances in neural information processing systems}, 36, 2024{\natexlab{b}}.

\bibitem[Lu et~al.(2024)Lu, Liu, Zhang, Wang, Dong, Liu, Sun, Ren, Li, Yang, et~al.]{lu2024deepseek}
Haoyu Lu, Wen Liu, Bo Zhang, Bingxuan Wang, Kai Dong, Bo Liu, Jingxiang Sun, Tongzheng Ren, Zhuoshu Li, Hao Yang, et~al.
\newblock Deepseek-vl: towards real-world vision-language understanding.
\newblock \emph{arXiv preprint arXiv:2403.05525}, 2024.

\bibitem[Moon et~al.(2022)Moon, Lee, Shin, Kim, and Choi]{moon2022multi}
Jong~Hak Moon, Hyungyung Lee, Woncheol Shin, Young-Hak Kim, and Edward Choi.
\newblock Multi-modal understanding and generation for medical images and text via vision-language pre-training.
\newblock \emph{IEEE Journal of Biomedical and Health Informatics}, 26\penalty0 (12):\penalty0 6070--6080, 2022.

\bibitem[Radford et~al.(2021)Radford, Kim, Hallacy, Ramesh, Goh, Agarwal, Sastry, Askell, Mishkin, Clark, et~al.]{radford2021learning}
Alec Radford, Jong~Wook Kim, Chris Hallacy, Aditya Ramesh, Gabriel Goh, Sandhini Agarwal, Girish Sastry, Amanda Askell, Pamela Mishkin, Jack Clark, et~al.
\newblock Learning transferable visual models from natural language supervision.
\newblock In \emph{International conference on machine learning}, pages 8748--8763. PMLR, 2021.

\bibitem[Ramesh et~al.(2021)Ramesh, Pavlov, Goh, Gray, Voss, Radford, Chen, and Sutskever]{ramesh2021zero}
Aditya Ramesh, Mikhail Pavlov, Gabriel Goh, Scott Gray, Chelsea Voss, Alec Radford, Mark Chen, and Ilya Sutskever.
\newblock Zero-shot text-to-image generation.
\newblock In \emph{International conference on machine learning}, pages 8821--8831. Pmlr, 2021.

\bibitem[Thawkar et~al.(2023)Thawkar, Shaker, Mullappilly, Cholakkal, Anwer, Khan, Laaksonen, and Khan]{thawkar2023xraygpt}
Omkar Thawkar, Abdelrahman Shaker, Sahal~Shaji Mullappilly, Hisham Cholakkal, Rao~Muhammad Anwer, Salman Khan, Jorma Laaksonen, and Fahad~Shahbaz Khan.
\newblock Xraygpt: Chest radiographs summarization using medical vision-language models.
\newblock \emph{arXiv preprint arXiv:2306.07971}, 2023.

\bibitem[Touvron et~al.(2023)Touvron, Lavril, Izacard, Martinet, Lachaux, Lacroix, Rozi{\`e}re, Goyal, Hambro, Azhar, et~al.]{touvron2023llama}
Hugo Touvron, Thibaut Lavril, Gautier Izacard, Xavier Martinet, Marie-Anne Lachaux, Timoth{\'e}e Lacroix, Baptiste Rozi{\`e}re, Naman Goyal, Eric Hambro, Faisal Azhar, et~al.
\newblock Llama: Open and efficient foundation language models.
\newblock \emph{arXiv preprint arXiv:2302.13971}, 2023.

\bibitem[Wang et~al.(2024)Wang, Zhao, Marostica, Yuan, Jin, Zhang, Li, Tang, Wang, Li, et~al.]{wang2024pathology}
Xiyue Wang, Junhan Zhao, Eliana Marostica, Wei Yuan, Jietian Jin, Jiayu Zhang, Ruijiang Li, Hongping Tang, Kanran Wang, Yu Li, et~al.
\newblock A pathology foundation model for cancer diagnosis and prognosis prediction.
\newblock \emph{Nature}, pages 1--9, 2024.

\bibitem[Wu et~al.(2023)Wu, Zhang, Zhang, Wang, and Xie]{wu2023towards}
Chaoyi Wu, Xiaoman Zhang, Ya Zhang, Yanfeng Wang, and Weidi Xie.
\newblock Towards generalist foundation model for radiology.
\newblock \emph{arXiv preprint arXiv:2308.02463}, 2023.

\bibitem[Wu et~al.(2024)Wu, Lin, Zhang, Zhang, Xie, and Wang]{wu2024pmc}
Chaoyi Wu, Weixiong Lin, Xiaoman Zhang, Ya Zhang, Weidi Xie, and Yanfeng Wang.
\newblock Pmc-llama: toward building open-source language models for medicine.
\newblock \emph{Journal of the American Medical Informatics Association}, page ocae045, 2024.

\bibitem[Wu et~al.(2021)Wu, Agu, Lourentzou, Sharma, Paguio, Yao, Dee, Mitchell, Kashyap, Giovannini, et~al.]{wu2021chest}
Joy~T Wu, Nkechinyere~N Agu, Ismini Lourentzou, Arjun Sharma, Joseph~A Paguio, Jasper~S Yao, Edward~C Dee, William Mitchell, Satyananda Kashyap, Andrea Giovannini, et~al.
\newblock Chest imagenome dataset for clinical reasoning.
\newblock \emph{arXiv preprint arXiv:2108.00316}, 2021.

\bibitem[Xu et~al.(2023{\natexlab{a}})Xu, Liu, Khan, Fan, and Wu]{xu2023multi}
Li Xu, Bo Liu, Ameer~Hamza Khan, Lu Fan, and Xiao-Ming Wu.
\newblock Multi-modal pre-training for medical vision-language understanding and generation: An empirical study with a new benchmark.
\newblock \emph{arXiv preprint arXiv:2306.06494}, 2023{\natexlab{a}}.

\bibitem[Xu et~al.(2023{\natexlab{b}})Xu, Shao, Zhang, Gao, Liu, Lei, Meng, Huang, Qiao, and Luo]{xu2023lvlm}
Peng Xu, Wenqi Shao, Kaipeng Zhang, Peng Gao, Shuo Liu, Meng Lei, Fanqing Meng, Siyuan Huang, Yu Qiao, and Ping Luo.
\newblock Lvlm-ehub: A comprehensive evaluation benchmark for large vision-language models.
\newblock \emph{arXiv preprint arXiv:2306.09265}, 2023{\natexlab{b}}.

\bibitem[Ye et~al.(2023)Ye, Xu, Xu, Ye, Yan, Zhou, Wang, Hu, Shi, Shi, et~al.]{ye2023mplug}
Qinghao Ye, Haiyang Xu, Guohai Xu, Jiabo Ye, Ming Yan, Yiyang Zhou, Junyang Wang, Anwen Hu, Pengcheng Shi, Yaya Shi, et~al.
\newblock mplug-owl: Modularization empowers large language models with multimodality.
\newblock \emph{arXiv preprint arXiv:2304.14178}, 2023.

\bibitem[Zhang et~al.(2023{\natexlab{a}})Zhang, Xu, Usuyama, Xu, Bagga, Tinn, Preston, Rao, Wei, Valluri, et~al.]{zhang2023biomedclip}
Sheng Zhang, Yanbo Xu, Naoto Usuyama, Hanwen Xu, Jaspreet Bagga, Robert Tinn, Sam Preston, Rajesh Rao, Mu Wei, Naveen Valluri, et~al.
\newblock Biomedclip: a multimodal biomedical foundation model pretrained from fifteen million scientific image-text pairs.
\newblock \emph{arXiv preprint arXiv:2303.00915}, 2023{\natexlab{a}}.

\bibitem[Zhang et~al.(2019)Zhang, Kishore, Wu, Weinberger, and Artzi]{zhang2019bertscore}
Tianyi Zhang, Varsha Kishore, Felix Wu, Kilian~Q Weinberger, and Yoav Artzi.
\newblock Bertscore: Evaluating text generation with bert.
\newblock \emph{arXiv preprint arXiv:1904.09675}, 2019.

\bibitem[Zhang et~al.(2023{\natexlab{b}})Zhang, Wu, Zhao, Lin, Zhang, Wang, and Xie]{zhang2023pmc}
Xiaoman Zhang, Chaoyi Wu, Ziheng Zhao, Weixiong Lin, Ya Zhang, Yanfeng Wang, and Weidi Xie.
\newblock Pmc-vqa: Visual instruction tuning for medical visual question answering.
\newblock \emph{arXiv preprint arXiv:2305.10415}, 2023{\natexlab{b}}.

\bibitem[Zhang et~al.(2024)Zhang, Wu, Zhao, Lei, Zhang, Wang, and Xie]{zhang2024radgenome}
Xiaoman Zhang, Chaoyi Wu, Ziheng Zhao, Jiayu Lei, Ya Zhang, Yanfeng Wang, and Weidi Xie.
\newblock Radgenome-chest ct: A grounded vision-language dataset for chest ct analysis.
\newblock \emph{arXiv preprint arXiv:2404.16754}, 2024.

\bibitem[Zhao et~al.(2024{\natexlab{a}})Zhao, Liu, Liu, Shi, and Wu]{zhao2024easygen}
Xiangyu Zhao, Bo Liu, Qijiong Liu, Guangyuan Shi, and Xiao-Ming Wu.
\newblock Easygen: Easing multimodal generation with bidiffuser and llms.
\newblock In \emph{Proceedings of the 62nd Annual Meeting of the Association for Computational Linguistics (Volume 1: Long Papers)}, pages 1351--1370, 2024{\natexlab{a}}.

\bibitem[Zhao et~al.(2024{\natexlab{b}})Zhao, Wang, Gu, Zhu, Mei, Zhuang, Cui, Wang, and Shen]{zhao2024chatcad}
Zihao Zhao, Sheng Wang, Jinchen Gu, Yitao Zhu, Lanzhuju Mei, Zixu Zhuang, Zhiming Cui, Qian Wang, and Dinggang Shen.
\newblock Chatcad+: Towards a universal and reliable interactive cad using llms.
\newblock \emph{IEEE Transactions on Medical Imaging}, 2024{\natexlab{b}}.

\bibitem[Zhou et~al.(2023)Zhou, He, Sun, Xu, Chen, Chu, Zhou, Liao, Zhang, and Gao]{zhou2023skingpt}
Juexiao Zhou, Xiaonan He, Liyuan Sun, Jiannan Xu, Xiuying Chen, Yuetan Chu, Longxi Zhou, Xingyu Liao, Bin Zhang, and Xin Gao.
\newblock Skingpt-4: an interactive dermatology diagnostic system with visual large language model.
\newblock \emph{arXiv preprint arXiv:2304.10691}, 2023.

\bibitem[Zhu et~al.(2023)Zhu, Chen, Shen, Li, and Elhoseiny]{zhu2023minigpt}
Deyao Zhu, Jun Chen, Xiaoqian Shen, Xiang Li, and Mohamed Elhoseiny.
\newblock Minigpt-4: Enhancing vision-language understanding with advanced large language models.
\newblock \emph{arXiv preprint arXiv:2304.10592}, 2023.

\bibitem[Zou et~al.(2024)Zou, Bai, Chen, Zhou, Chen, Ren, Wang, Yuan, Shen, and Fu]{zou2024medrg}
Ke Zou, Yang Bai, Zhihao Chen, Yang Zhou, Yidi Chen, Kai Ren, Meng Wang, Xuedong Yuan, Xiaojing Shen, and Huazhu Fu.
\newblock Medrg: Medical report grounding with multi-modal large language model.
\newblock \emph{arXiv preprint arXiv:2404.06798}, 2024.

\end{thebibliography}
}
\clearpage
\clearpage
\setcounter{page}{1}
\maketitlesupplementary

\begin{table}[!h]
\centering
\resizebox{\linewidth}{!}{
\begin{tabular}{p{0.5cm}|p{3.5cm}|p{3.5cm}}
\hline
& \textbf{Chest ImaGenome} & \textbf{Ours} \\
\hline
\multirow{8}{*}{\rotatebox{90}{Reserve}} & right lung, right mid lung zone, right hilar structures, right hemidiaphragm, left lung, left mid lung zone, left hilar structures, left hemidiaphragm, trachea, spine, abdomen, svc& right lung, right mid lung zone, right hilar structures, right hemidiaphragm, left lung, left mid lung zone, left hilar structures, left hemidiaphragm, trachea, spine, abdomen, svc\\

\hline
\multirow{13}{*}{\rotatebox{90}{Incorporate}}&left upper lung zone& \multirow{2}{*}{left upper lung zone}\\
& left apical zone& \\
\cline{2-3}
&right upper lung zone& \multirow{2}{*}{right upper lung zone}\\
&right apical zone& \\
\cline{2-3}
&mediastinum& \multirow{2}{*}{mediastinum}\\
&upper mediastinum& \\ 
\cline{2-3}
&right lower lung zone& \multirow{2}{*}{right lower lung zone}\\
&right costophrenic angle& \\
\cline{2-3}
&left lower lung zone& \multirow{2}{*}{left lower lung zone}\\
&left costophrenic angle& \\
\cline{2-3}
&cardiac silhouette& \multirow{3}{*}{cardiac silhouette}\\ 
&cavoatrial junction& \\
&right atrium& \\

\hline
\multirow{4}{*}{\rotatebox{90}{Delete}}&carina&  \multirow{4}{*}{-}\\
&right clavicle& \\
&left clavicle& \\
&aortic arch& \\
\hline
\multirow{19}{*}{\rotatebox{90}{Merge}} & left lung + right lung & bilateral lung\\
\cline{2-3}
& left upper + right upper & bilateral upper lung zone  \\
\cline{2-3}
& left mid + right mid & bilateral mid lung zone \\
\cline{2-3}
& left lower  + right lower & bilateral lower lung zone \\ 
\cline{2-3}
& left hilar + right hilar & bilateral hilar structures\\
\cline{2-3}
& left hemidiaphragm + right hemidiaphragm & bilateral hemidiaphragm \\
\cline{2-3}
& left mid + left lower & left mid-to-lower lung zone \\
\cline{2-3}
& right mid + right lower & right mid-to-lower lung zone\\
\cline{2-3}
& left mid + left upper & left mid-to-upper lung zone \\
\cline{2-3}
& right mid + right upper & right mid-to-upper lung zone \\
\cline{2-3}
& left mid-to-lower + right mid-to-lower & bilateral mid-to-lower lung zone\\
\cline{2-3}
& left mid-to-upper + right mid-to-upper & bilateral mid-to-upper lung zone\\
\hline
Sum & 29 & 30\\

\end{tabular}
}
\caption{Anatomical regions transformation from the Chest ImaGenome to our GEMeX version. The left column indicates the detailed operation.}
\label{tab:anatomical_regions}
\end{table}

\section{GEMeX Construction Details}

\subsection{Transformation and Distribution of Anatomical Regions}
\begin{figure}[!h]
  \centering
  \includegraphics[width=1.0\linewidth]{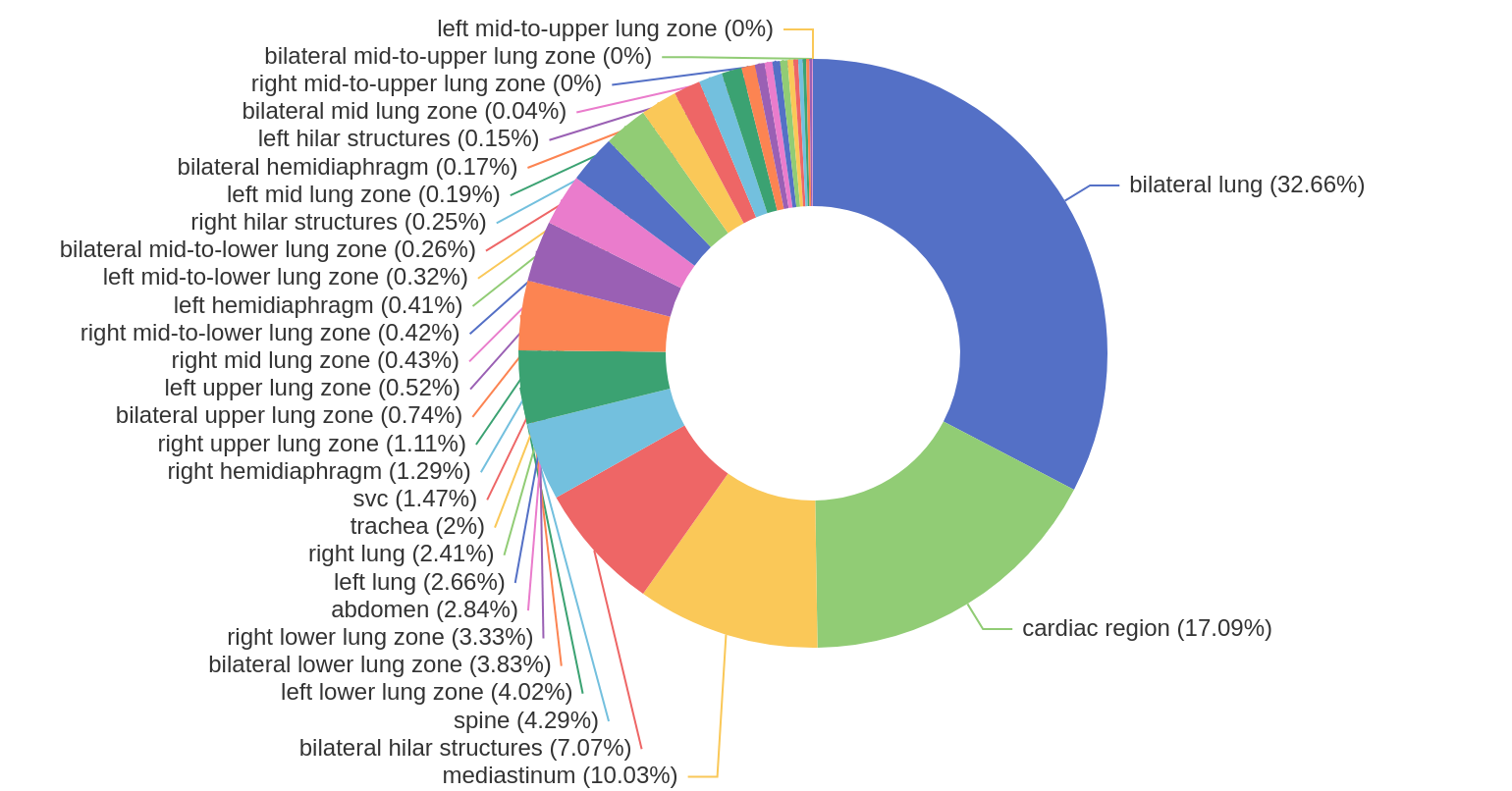}
  \caption{
  Distribution of anatomical regions corresponding to each sentence after transformation from the Chest ImaGenome dataset.
  }
  \label{fig:pie_region}
\end{figure}

As we said in Section~\ref{sec:region_selection_merge}, we provide detailed operations to transform anatomical regions from Chest ImaGenome to our \shortname{}. The process is summarized in Table~\ref{tab:anatomical_regions}. 
The resulting anatomical region distribution corresponding to each sentence is shown in Figure~\ref{fig:pie_region}. Overall, there are 30 regions, and the merged area occupies a large proportion, such as ``bilateral lung'' and ``bilateral hilar structures''.

\vspace{1cm}

\begin{table*}[t!]\centering
\begin{minipage}{2.0\columnwidth}\vspace{0mm}    \centering
\begin{tcolorbox} 
    \centering
      \footnotesize
\begin{tabular}{p{0.97\columnwidth} c}

{\bf messages} = [``role'':``{\bf system}'', ``content'': \\
f```You are a helpful chest X-ray radiologist. Given an input sentence, your task is to map it to an anatomical region on X-ray for better observation from 
a predefined list [right lung, cardiac silhouette,..., bilateral lower lung zone]. \\ 

\underline{Here are some rules}:\\
\emph{(1)} If there is no corresponding region for this sentence, leave it out.\\
\emph{(2)} If the sentence describes the overall anatomical characteristics without specifying a particular region, you can choose ``bilateral lung'' as its region. For example, ``No focal consolidation, pleural effusion or pneumothorax is present'':``bilateral lung''. \\
\emph{(3)} One sentence can only correspond to one region. If a sentence's main symptom involves several anatomical regions, rephrase it into multiple sentences with corresponding regions. Note that all derived sentences must be syntactically complete, not phrases (i.e., containing a subject and a predicate at least). For example: ``The cardiomediastinal silhouette is normal.'' can be segmented into {``The cardiac silhouette is normal.'':``cardiac region'', ``The mediastinal silhouette is normal.'':``mediastinum''}, where ``cardiomediastinal'' corresponds to the ``cardiac region'' and ``mediastinum''.\\
\emph{(4)} Small (tiny) pleural effusion (fluid) usually appears in the lower lung zone, a moderate pleural effusion appears in the mid-to-lower lung zone, and a large (substantial) pleural effusion can even occupy the entire lung. If the severity (like small, moderate and large) is not indicated, output the left lung or right lung directly.\\
\emph{(5)} The main anatomical region for observing pulmonary venous is the bilateral hilar structures on the X-ray.\\
\emph{(6)} The region where the atrium and ventricle can be observed is the cardiac region.\\
\underline{Here are some cases}:
(1)... (2)... (3)... (4)... \\
Organize your output in a json formatted as Dict{Str(sentence):Str(region)}, without other words.''']\\

\hrulefill & \\
{\bf messages} += [``role'':``{\bf user}'', ``content'': ``Input: ``Bibasilar atelectasis is seen without discrete focal consolidation.''\\
\end{tabular}
\end{tcolorbox}
\vspace{-2mm}
\caption{Our proposed prompt guided by radiologist feedback for refining sentence-region pairs.}
    \label{tab:prompt_refinement}
\end{minipage}
\end{table*}

\begin{table*}[t!]\centering
\begin{minipage}{2.0\columnwidth}\vspace{0mm}    \centering
\begin{tcolorbox} 
    \centering
      \footnotesize
\begin{tabular}{p{0.9\columnwidth} c}

{\bf messages} = [``role'':``{\bf system}'', ``content'': \\
f```You are a chest X-ray AI assistant, and you are seeing a frontal view chest X-ray image, described by several phrases with visual regions.
                        Generate {\bf 3} open-ended questions, {\bf 2} closed-ended questions,  {\bf 3} single-choice questions, and {\bf 3} multi-choice questions about this chest X-ray. Format your output in JSON format. \\
\underline{Here are some rules}:\\
\emph{(1)} Include questions asking about the visual content of the image, containing abnormality, disease, location, severity, cause of disease, size, and implication. For a CXR, the types of questions generated need to be diverse. Do not ask any questions that cannot be answered confidently.\\
\emph{(2)} For each question, generate its type (abnormality, location, ...), provide the answer, explain the reason for obtaining such answer, and output the corresponding visual regions as a visual clue.\\
\emph{(3)} For open-ended questions, the answers must be concise. You should generate detailed reasons based on the provided CXR phrases and your medical knowledge. Do not refer to the text description in your questions or answers. \\
\emph{(4)} Avoid questions that cannot be answered by looking at the given CXR image itself, such as asking about changes/comparisons from previous scans, asking about staff notifications, or asking about view types or other scans. \\ 
\underline{Here is one example}:\\
Chest X-ray: \{...\}, One open-ended question can be: \{...\}, One closed-ended question can be: \{...\}, One single-choice question can be: \{...\}, One multi-choice question can be: \{...\} \\

\hrulefill & \\
{\bf messages} += [``role'':``{\bf user}'', ``content'': ``Chest X-ray: There is also fullness of the right hilum which is new. [visual location: right hilar structures] ...''\\

\end{tabular}
\end{tcolorbox}
\vspace{-2mm}
\caption{Our designed prompt for generating groundable and explainable medical VQA, using a grounded report as input.}
\label{tab:prompt_vqa_generation}
\end{minipage}
\end{table*}

\subsection{Prompt for Re-grounding Report}
Here, we provide detailed instructions to re-ground reports with medical LLM, as we elaborated in Section~\ref{sec:report_refinement}. The prompt is shown in Table~\ref{tab:prompt_refinement}, where we add clinical guidance (like (2) (4) (5) (6)) and split and re-written requirements (\eg, (3)) to ensure correct sentence-region correspondence. Moreover, we provide some manually labeled pairs as demonstrations (\ie, ``here are some cases'') for in-context learning, aiming to improve overall performance.

\subsection{Prompt for VQA Generation}
In Table~\ref{tab:prompt_vqa_generation}, we provide a detailed prompt to guide GPT-4o in VQA generation. Specifically, for each CXR, we generate 3 open-ended questions, 2 closed-ended questions, 3 single-choice questions, and 3 multi-choice
questions, containing diverse content, like ``abnormality'', ``disease'', etc. GPT-4o is required to provide detailed reasoning and specify relevant visual regions  (rule (2)). Additionally, comparison-based questions are excluded, as only a single CXR is provided (rule (3)). To better align with our objectives, we incorporate manually crafted questions as demonstrations (the inputs after ``Here is one example'').

\begin{figure*}[t]
    \begin{subfigure}[b]{0.32\textwidth}
        \centering
        \includegraphics[width=\textwidth]{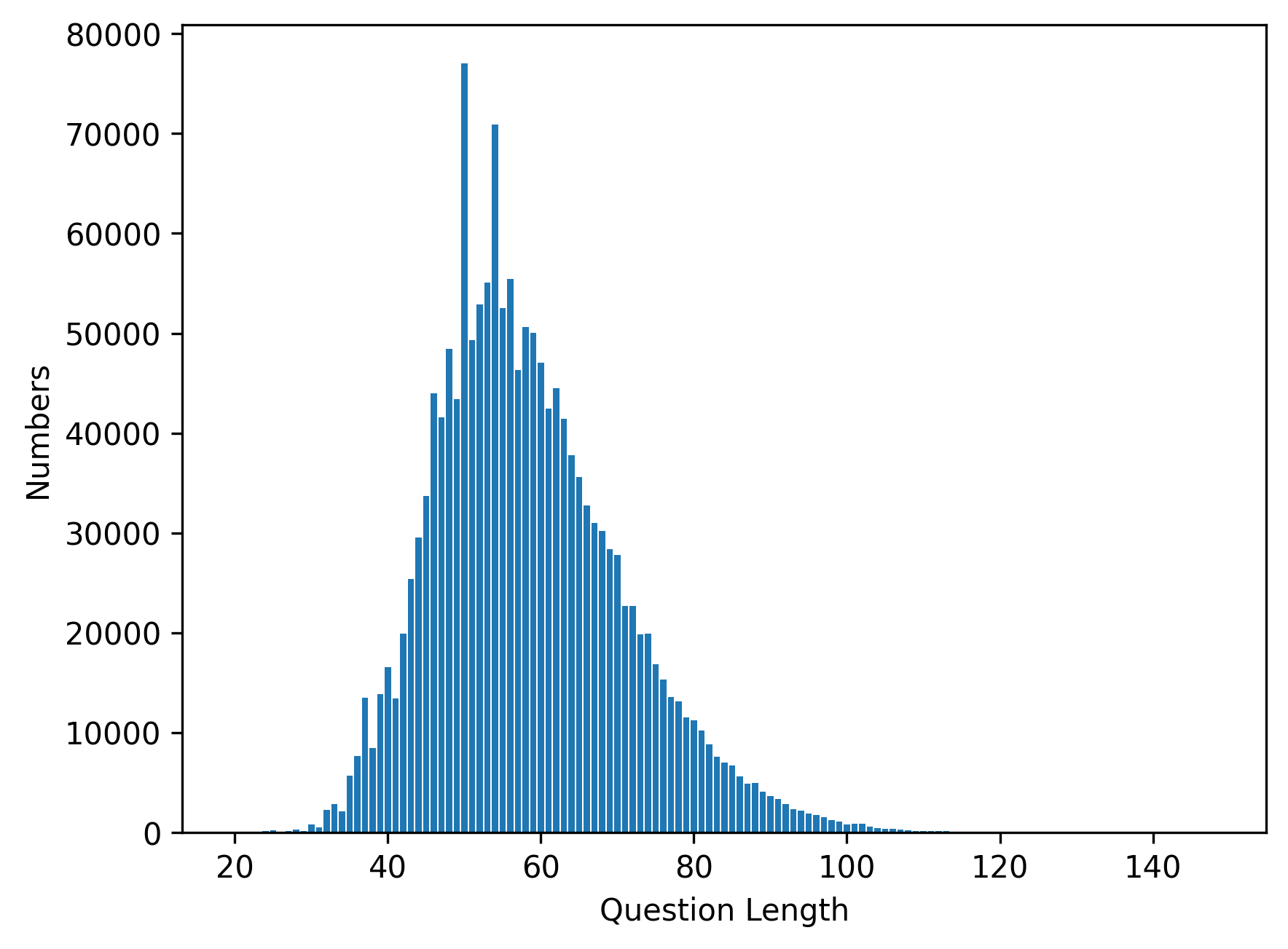}
        \caption{Distribution of question lengths in \underline{training} set.}
    \end{subfigure}
    \hfill
    \begin{subfigure}[b]{0.32\textwidth}
        \centering
        \includegraphics[width=\textwidth]{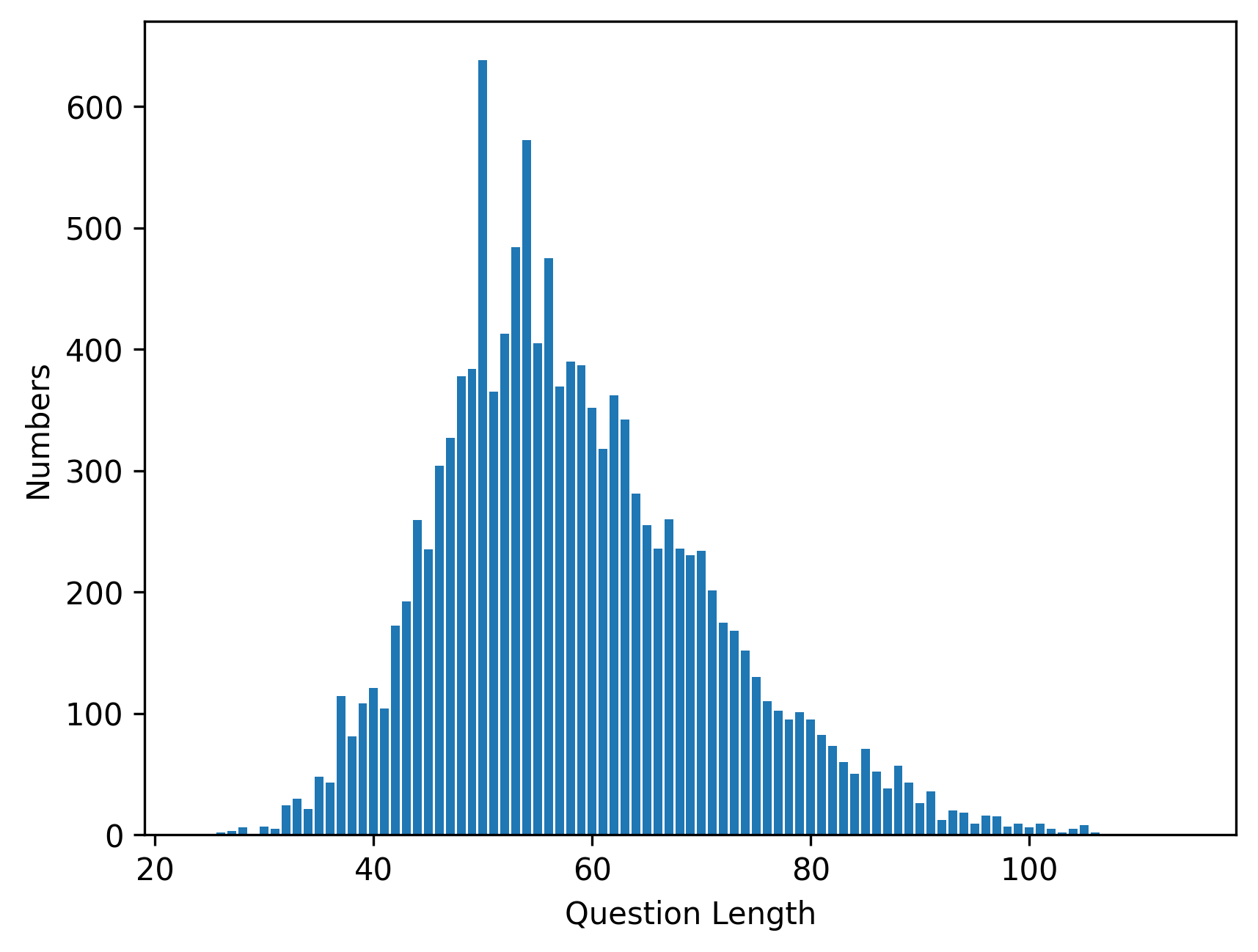}
        \caption{Distribution of question lengths in \underline{valid} set.}
    \end{subfigure}
    \hfill
    \begin{subfigure}[b]{0.32\textwidth}
        \centering
        \includegraphics[width=\textwidth]{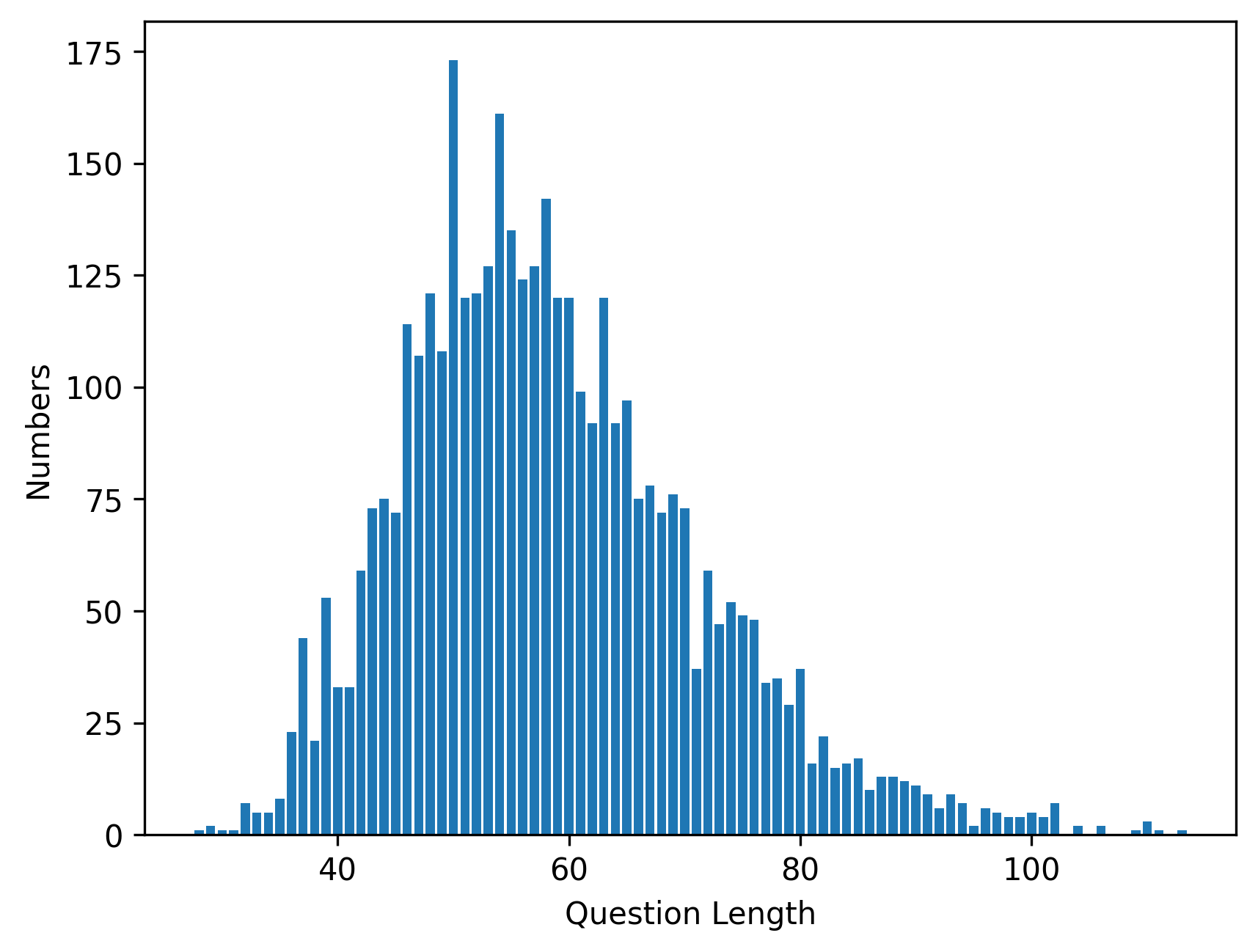}
        \caption{Distribution of question lengths in \underline{test} set.}
    \end{subfigure}
    
    \vspace{0.5cm}
    \begin{subfigure}[b]{0.32\textwidth}
        \centering
        \includegraphics[width=\textwidth]{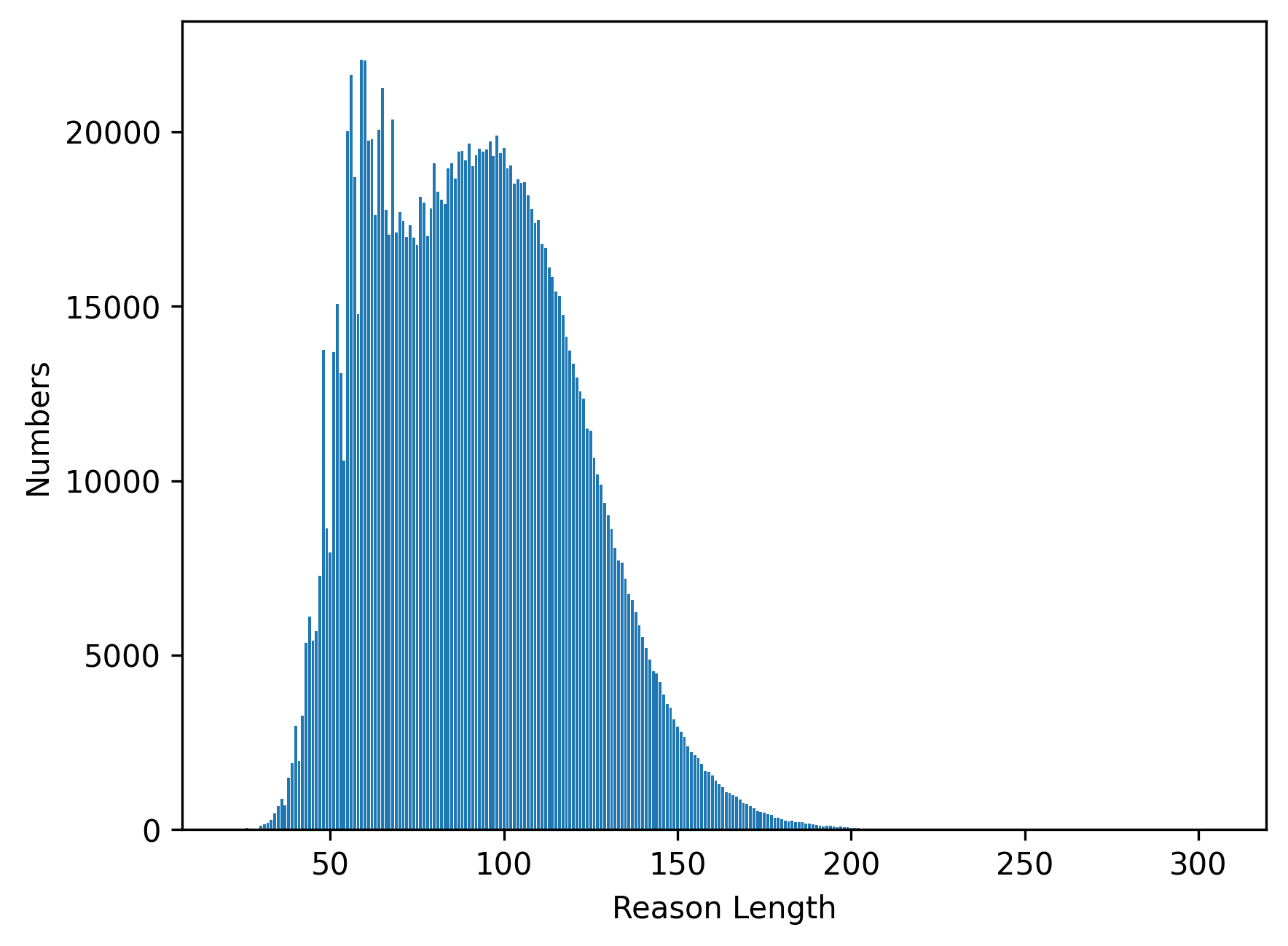}
        \caption{Distribution of reason lengths in \underline{training} set.}
    \end{subfigure}
    \hfill
    \begin{subfigure}[b]{0.32\textwidth}
        \centering
        \includegraphics[width=\textwidth]{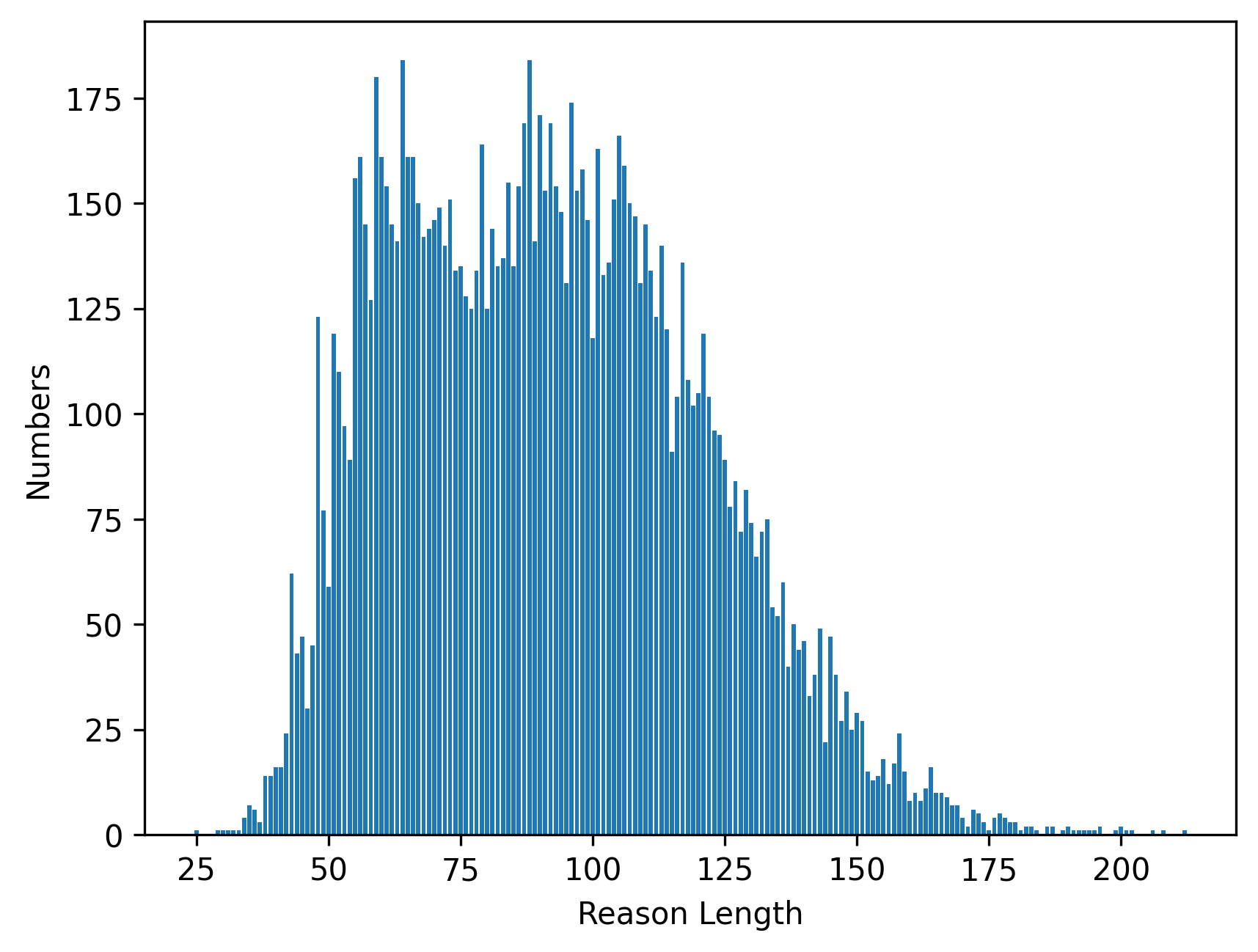}
        \caption{Distribution of reason lengths in \underline{valid} set.}
    \end{subfigure}
    \hfill
    \begin{subfigure}[b]{0.32\textwidth}
        \centering
        \includegraphics[width=\textwidth]{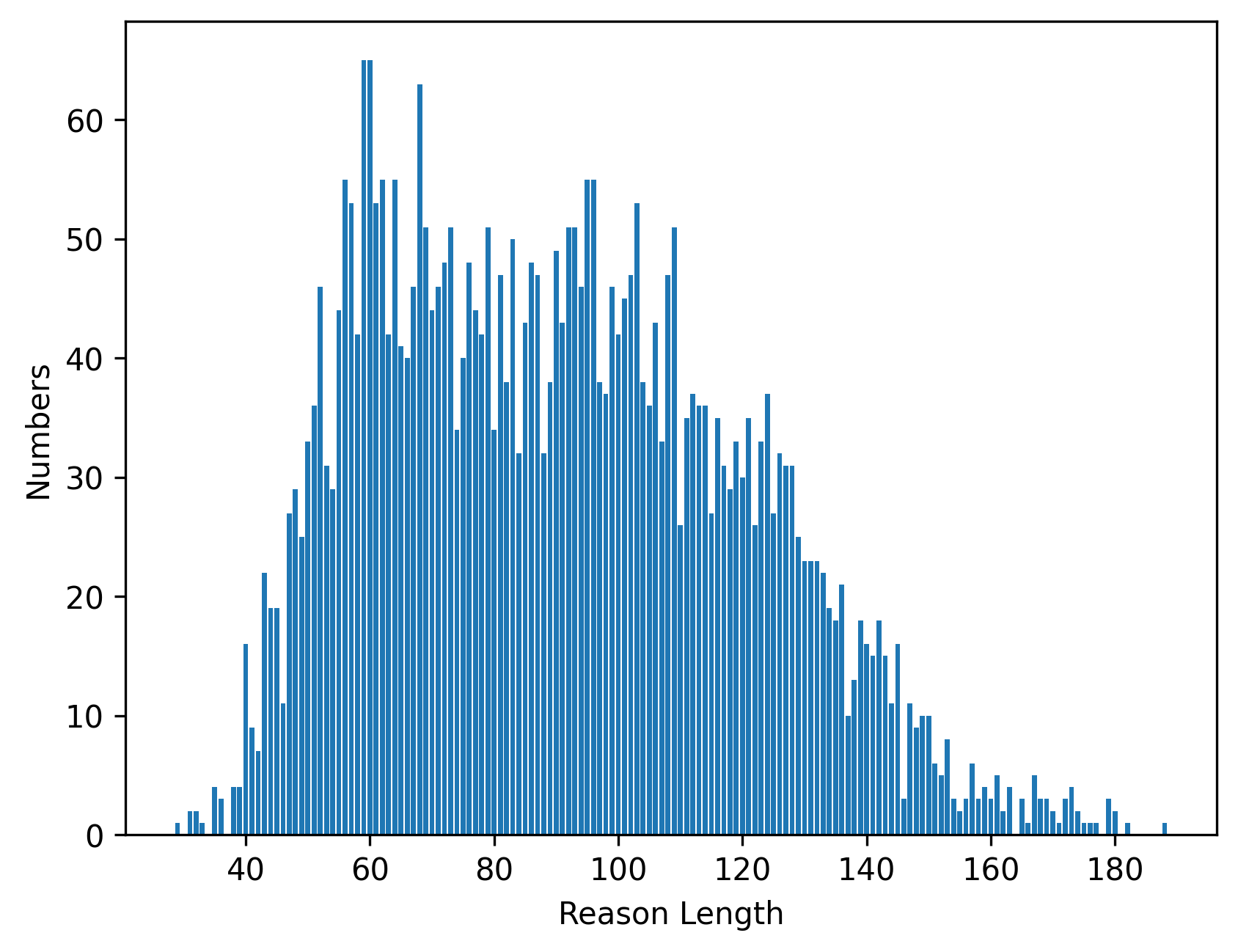}
        \caption{Distribution of reason lengths in \underline{test} set.}
    \end{subfigure}
    
    \caption{Detailed distribution of question lengths and reason lengths across data splits.}
    \label{fig:length_distribution}
\end{figure*}

\begin{figure*}[t]
    \centering
    \begin{subfigure}[b]{0.48\textwidth}
        \centering
        \includegraphics[width=\textwidth]{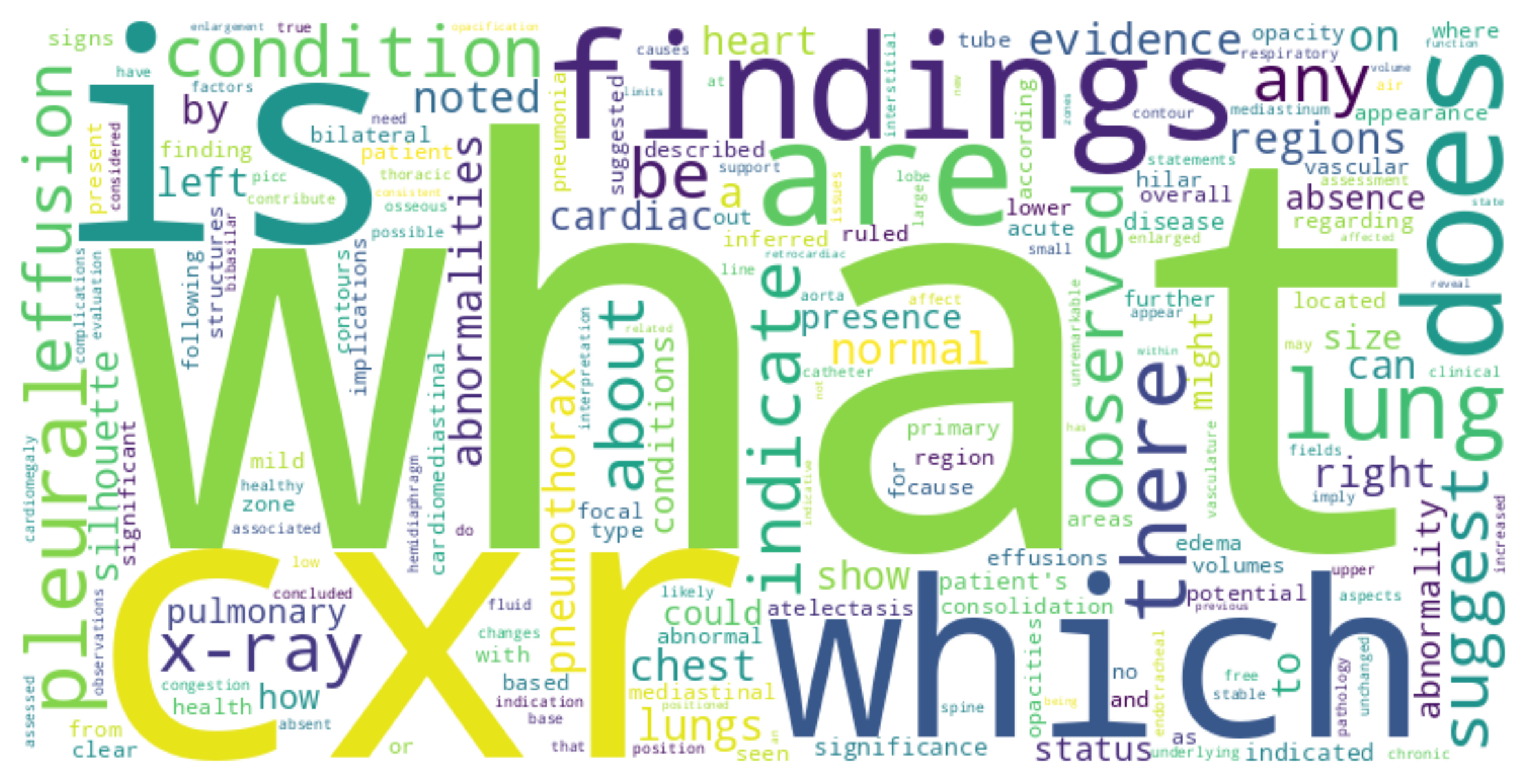}
        \label{fig:image1}
    \end{subfigure}
    \begin{subfigure}[b]{0.48\textwidth}
        \centering
        \includegraphics[width=\textwidth]{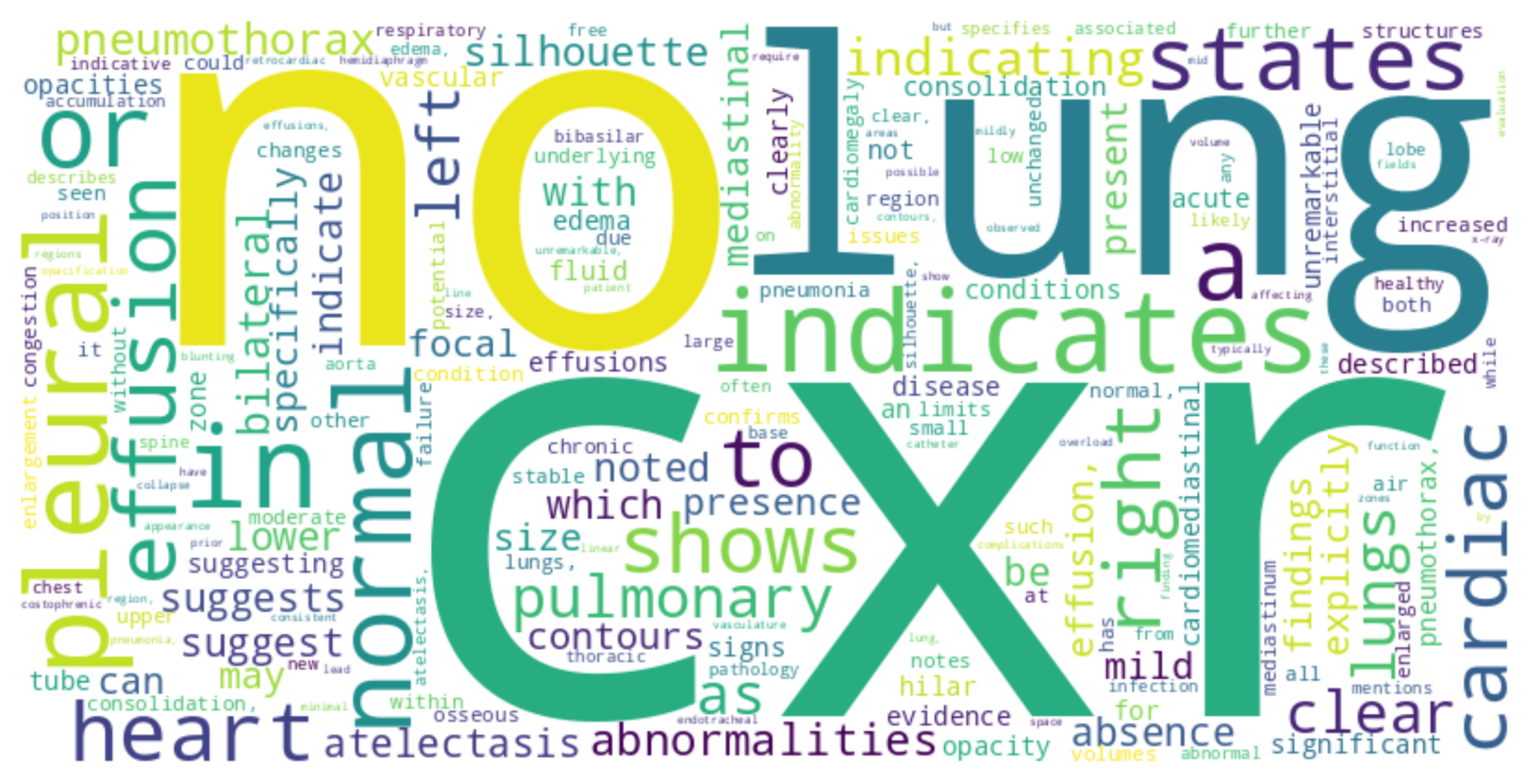}
        \label{fig:image2}
    \end{subfigure}
    \caption{Visualization of word frequencies in questions (left) and reasons (right).}
    \label{fig:word_cloud}
\end{figure*}



\section{Benchmark Details}

\subsection{Distribution Analysis of Question and Reason Lengths Across Data Splits}
We provide a detailed distribution of question and reason lengths across data splits, as presented in Figure~\ref{fig:length_distribution}. It can be seen that under different data splits, the distributions of question (reason) lengths are generally similar. Furthermore, from the perspective of reason, the reason lengths mostly fall between 60 and 150, demonstrating the level of detail in the reasons as textual explanations.

\subsection{Word Frequency Analysis of Questions and Reasons}
Besides providing length distribution, we also explore the frequency of words from both questions and reasons, as shown in Figure~\ref{fig:word_cloud}. From the left part (regarding questions), we can observe that the majority of words are question-related terms, such as ``what'', ``which'', ``is'', and ``are''. Additionally, some disease-related terms, such as ``abnormality'', ``findings'', and ``pleural effusion'', are also quite common. Lastly, content words related to the questions, such as ``regions'', ``evidence'', and ``size'', are frequently mentioned. These demonstrate the diversity of questions; On the right, we show the word cloud of reasons. It can be seen that the vocabulary mainly falls into two categories: one is related to diseases or anatomical regions, such as ``normal'', ``heart'', and ``pleural'', and the other consists of words used to convey explanations, such as ``indicates'' and ``states''.

\subsection{Input Samples for the Proposed Fine-tuning}
We here provide some input samples for a better understanding of how to fine-tune LLaVA-Med~\cite{li2024llava} on our \shortname{}. An input sample of open-ended questions can be seen in Table~\ref{tab:input_open_ended}; An input sample of closed-ended questions can be seen in Table~\ref{tab:input_closed_ended}; An input sample of single-choice questions can be seen in Table~\ref{tab:input_single_choice}; An input sample of multi-choice questions can be seen in Table~\ref{tab:input_multi_choice}.

\subsection{Fine-tuning Details}
We fine-tune both the visual projection layers and the LLM components of LLaVA-Med-v1 (after stage II) by calculating the auto-regressive loss to predict the assistant’s responses and the dialogue termination token \texttt{<STOP>}. Particularly, the model is trained for 3 epochs on four NVIDIA H100 GPUs with a batch size of 64, taking around 54 hours. The network is warmed up in the ﬁrst 0.03 epochs with a linear learning rate from 3e-7 to 2e-5, which further decays by cosine schedule.
The optimizer is AdamW. To accelerate training, we employ the Fully Sharded Data Parallel (FSDP) mechanism, the bf16 (Brain Floating Point) data format, and gradient checkpointing.

\subsection{LVLMs Introduction}
Besides fine-tuning a task-oriented model, we perform a zero-shot evaluation on our \shortname{} dataset across the other 12 LVLMs, with 7 in the general domain and the other 5 in the medical domain:
\begin{itemize}
    \item \textbf{In the General Domain:} \texttt{LLaVA-v1}~\cite{liu2024visual} and \texttt{Mini-GPT4-v1}~\cite{zhu2023minigpt} are two pioneering works, achieving remarkable results in multimodal tasks; \texttt{mPLUG-Owl}~\cite{ye2023mplug} is a multimodal model from the mPLUG series focused on visual-language tasks; \texttt{LLaVA-v1.5}~\cite{liu2024improved} is an improved version of LLaVA-v1 with enhancements in multimodal alignment, accuracy, and efficiency; 
    \texttt{Deepseek-VL}~\cite{lu2024deepseek} uses the SigLIP-L and SAM-B as the hybrid vision encoder and the DeepSeek-LLM as the base LLM for solving multimodal tasks;
    \texttt{Qwen-VL-Chat}~\cite{bai2023qwen} originates from the Qwen family, possessing capabilities such as multilingual dialogue and multi-image interleaved dialogue;
    \texttt{GPT-4o-mini} is a smaller, optimized version of GPT-4~\cite{achiam2023gpt} intended for lighter computational environments. Note that we did not test GPT-4o because its safety protection policy prohibits it from analyzing medical images.

    \item \textbf{In the Medical Domain:} \texttt{LLaVA-Med-v1}~\cite{li2024llava} is designed specifically for medical applications based on LLaVA-v1 and \texttt{LLaVA-Med-v1.5}~\cite{li2024llava} is an advanced version; \texttt{MiniGPT-Med}~\cite{alkhaldi2024minigpt} is an medical version of Mini-GPT4~\cite{zhu2023minigpt}; \texttt{XrayGPT}~\cite{thawkar2023xraygpt} is a specialized GPT model for interpreting chest X-rays; \texttt{RadFM}~\cite{wu2023towards} is a radiology foundation model.

\end{itemize}

To make a fair comparison, the evaluated models (except GPT-4o-mini and RadFM (with MedLLaMA-13B~\cite{wu2024pmc})) are based on 7B-LLMs in this section. 
Specifically, LLaVA-v1, LLaVA-Med-v1, and Mini-GPT4-v1 are based on Vicuna-v0-7B~\cite{chiang2023vicuna} while LLaVA-v1.5 and XrayGPT are based on Vicuna-v1-7B; LLaVA-Med-v1.5 is built upon Mistral-7B-Instruct-v0.2~\cite{jiang2023mistral};  mPLUG-Owl is using LLaMA-7B~\cite{touvron2023llama}; Deepseek-VL~\cite{lu2024deepseek} is based on DeepSeek-LLM-7B; and Qwen-VL-Chat~\cite{bai2023qwen} is based on Qwen-7B.
All models' configurations are set according to their open-source codes.

\begin{table}[t!]\centering
\begin{minipage}{0.95\columnwidth}\vspace{0mm}    \centering
\begin{tcolorbox} 
    \centering
      \footnotesize
\begin{tabular}{p{0.97\columnwidth} c}
A chat between a curious human and an artificial intelligence assistant. 
The assistant gives helpful, detailed, and polite answers to the human's questions. Input an open-ended question, and the assistant will output its answer with a detailed reason and corresponding visual location.\textbackslash n\textbackslash n\#\#\# Human:\textless image\textgreater\textbackslash n What abnormalities are observed in the left lower lung zone?\textbackslash n\#\#\# Assistant: \textless answer\textgreater Linear atelectasis.  \textless reason\textgreater The CXR indicates that the lungs are clear except for linear atelectasis located specifically at the left base. \textless location\textgreater [[126, 110, 203, 167]]\textbackslash n\#\#\#
\end{tabular}
\end{tcolorbox}
\vspace{-2mm}
\caption{An input sample of open-ended questions}
\label{tab:input_open_ended}
\end{minipage}
\end{table}

\begin{table}[t!]\centering
\begin{minipage}{0.95\columnwidth}\vspace{0mm}    \centering
\begin{tcolorbox} 
    \centering
      \footnotesize
\begin{tabular}{p{0.97\columnwidth} c}
A chat between a curious human and an artificial intelligence assistant. 
The assistant gives helpful, detailed, and polite answers to the human's questions. Input a closed-ended question, and the assistant will output its answer (yes or no) with a detailed reason and corresponding visual location.\textbackslash n\textbackslash n\#\#\# Human:\textless image\textgreater\textbackslash n Are there any lung abnormalities present in this CXR?\textbackslash n\#\#\# Assistant: \textless answer\textgreater No.  \textless reason\textgreater The CXR clearly shows that the lungs are clear, indicating no lung abnormalities. \textless location\textgreater [[30, 34, 185, 178]]\textbackslash n\#\#\#
\end{tabular}
\end{tcolorbox}
\vspace{-2mm}
\caption{An input sample of closed-ended questions}
\label{tab:input_closed_ended}
\end{minipage}
\end{table}

\begin{table}[t!]\centering
\begin{minipage}{0.95\columnwidth}\vspace{0mm}    \centering
\begin{tcolorbox} 
    \centering
      \footnotesize
\begin{tabular}{p{0.97\columnwidth} c}
A chat between a curious human and an artificial intelligence assistant. 
The assistant gives helpful, detailed, and polite answers to the human's questions. Input a single-choice question, and the assistant will output its answer (an option) with a detailed reason and corresponding visual location.\textbackslash n\textbackslash n\#\#\# Human:\textless image\textgreater\textbackslash n Which of the following is absent in this CXR? \textless choices\textgreater: [A: Pulmonary edema, B: Pleural effusion, C: Pneumothorax, D: All of the above]\textbackslash n\#\#\# Assistant: \textless answer\textgreater D  \textless reason\textgreater The CXR shows that there is no pulmonary edema, effusion, or pneumothorax present. \textless location\textgreater [[48, 48, 175, 180]]\textbackslash n\#\#\#
\end{tabular}
\end{tcolorbox}
\vspace{-2mm}
\caption{An input sample of single-choice questions}
\label{tab:input_single_choice}
\end{minipage}
\end{table}

\begin{table}[t!]\centering
\begin{minipage}{0.95\columnwidth}\vspace{0mm}    \centering
\begin{tcolorbox} 
    \centering
      \footnotesize
\begin{tabular}{p{0.97\columnwidth} c}
A chat between a curious human and an artificial intelligence assistant. 
The assistant gives helpful, detailed, and polite answers to the human's questions. Input a multi-choice question, and the assistant will output its answer (some options) with a detailed reason and corresponding visual location.\textbackslash n\textbackslash n\#\#\# Human:\textless image\textgreater\textbackslash n What abnormalities are mentioned regarding the lung fields? \textless choices\textgreater: [A: Clear lung fields, B: Atelectasis, C: Effusions, D: Congestion]\textbackslash n\#\#\# Assistant: \textless answer\textgreater [B, C]  \textless reason\textgreater The CXR shows bibasilar atelectasis and small pleural effusions. There is no impressive congestion shown. \textless location\textgreater [[26, 119, 217, 183]]\textbackslash n\#\#\#
\end{tabular}
\end{tcolorbox}
\vspace{-2mm}
\caption{An input sample of multi-choice questions}
\label{tab:input_multi_choice}
\end{minipage}
\end{table}

\section{More Case Studies}
Here, we present more questions with answers from \texttt{GPT-4o-mini}, \texttt{LLaVA-Med}, and our fine-tuned \texttt{LLaVA-Med-\shortname{}}, for better understand our dataset \shortname{} and the corresponding performance of LVLMs. 
\begin{itemize}
    \item We first present some cases of \textbf{open-ended questions}, as in Table~\ref{tab:more_cases_open}. It can be seen that our fine-tuned model can generally provide correct (or partially correct) answers and identify relatively accurate visual locations. However, other models fail to deliver both precise textual answers and accurate visual positions simultaneously.
    \item Furthermore, we provide some cases from \textbf{closed-ended questions} in Table~\ref{tab:more_cases_closed}. Although LLaVA-Med can correctly answer the first two questions, it fails to provide visual grounding. For the third question, GPT-4o-mini provides a correct answer, but there is a discrepancy between its grounded visual location (mediastinum) and the ground truth (cardiac region). In contrast, our fine-tuned model can provide both correct answers and accurate visual grounding.
    \item Next, we show three cases of \textbf{single-choice questions}, presented in Table~\ref{tab:more_cases_single}. Overall, GPT-4o-mini and LLaVA-Med demonstrate insufficient image understanding capabilities. For instance, in the third example (CASE III), both models incorrectly identify pleural effusion on both sides, whereas it is actually present only in the left lower lung. In comparison, the fine-tuned model shows significant improvement in visual understanding, as evidenced by the grounding results.
    \item Finally, some cases from \textbf{multi-choice questions} are illustrated in Table~\ref{tab:more_cases_multi}. It can be observed that multi-choice questions are generally more challenging. GPT-4o-mini can only partially identify the correct options; LLaVA-Med tends to answer questions directly based on the question. For example, in all CASEs, it outputs all textually corresponding answers and provides seemingly reasonable explanations, but some of these answers are incorrect when judged based on visual content;  The fine-tuned model may also make analytical errors. In CASE III, it incorrectly determines the presence of pleural effusion.
\end{itemize}

\begin{table*}[t]
  \begin{minipage}{0.99\textwidth}
\centering  
\scalebox{0.70}{
\large
\begin{tabular}{l |p{6.5cm}| p{6.5cm} | p{6.5cm} }
\toprule
 \multicolumn{3}{l}{\bf Open-ended questions from \shortname{}:}  \\
\midrule
& \textbf{CASE I} & \textbf{CASE II} & \textbf{CASE III}\\
&  \includegraphics[height=4.5cm]{images/case_1.pdf} &  \includegraphics[height=4.5cm]{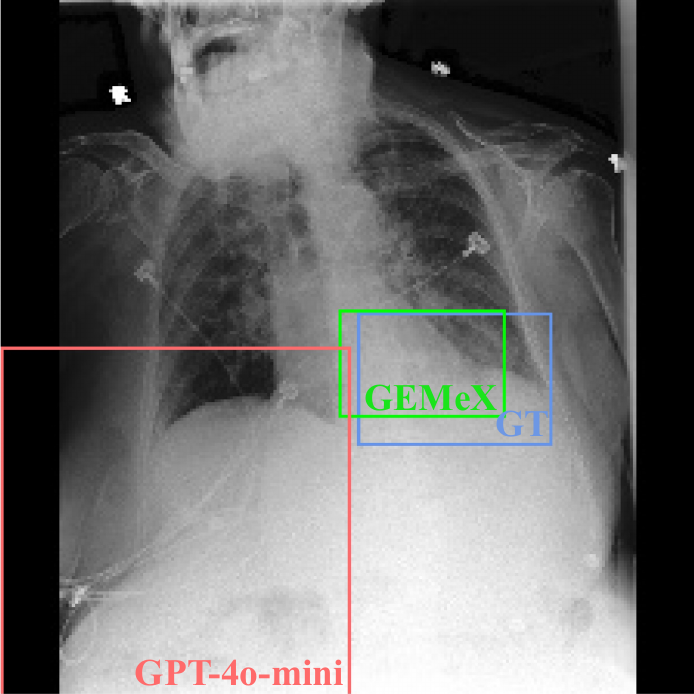} &
\includegraphics[height=4.5cm]{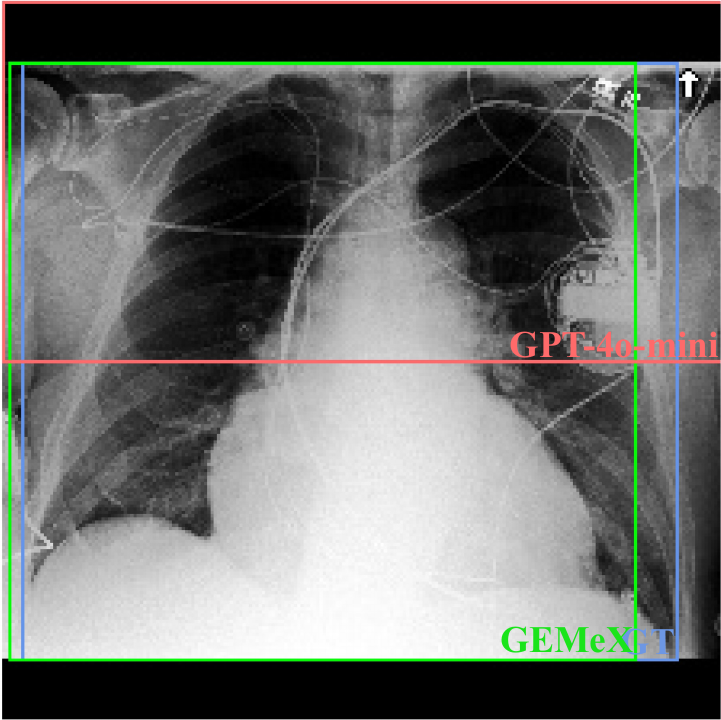}\\
& (Bbox predicted by each model) & (Bbox predicted by each model) &(Bbox predicted by each model)\\
 \midrule
\emph{Question} 
& What might be the cause of the overexpansion of the right lung?  
& What disease is indicated at the left lung base?
& What is the condition of the lungs in this CXR?
\\

\emph{Ground Truth} 
& \textless Answer\textgreater: \textcolor{gray}{Compensatory hyperinflation} due to left lung removal. \textless Reason\textgreater: After a pneumonectomy, the remaining lung may overexpand to compensate for the \textcolor{orange}{loss of the left lung}.  
&  \textless Answer\textgreater: \textcolor{gray}{Atelectasis and/or pulmonary edema}. \textless Reason\textgreater: The increased opacity at the left lung base suggests either atelectasis or a combination of atelectasis and pulmonary edema.
& \textless Answer\textgreater: Clear. \textless Reason\textgreater: The CXR indicates that \textcolor{gray}{the lungs appear clear} without definite signs of pneumonia or congestive heart failure (CHF).

\\ \midrule
\texttt{GPT-4o-mini} 
&  \textless Answer\textgreater: The overexpansion of the right lung may be caused by conditions such as a \textcolor{gray}{right-sided pneumothorax or chronic obstructive pulmonary disease (COPD) (\XSolidBrush)}.   \textless Reason\textgreater: Overexpansion can occur when air enters the pleural space (as in pneumothorax) or when there is an obstruction that prevents air from leaving the lung, causing it to become hyperinflated. In COPD, the lung tissue loses elasticity, leading to overinflation. \textless GPTScore\textgreater: 44.44 
& \textless Answer\textgreater: \textcolor{gray}{Possible pneumonia or pleural effusion (\XSolidBrush)}. \textless Reason\textgreater: The left lung base shows increased opacity, which may indicate fluid accumulation (pleural effusion) or consolidation (pneumonia). These conditions often present as areas of increased density on chest X-rays. \textless GPTScore\textgreater: 55.56
&  \textless Answer\textgreater: \textcolor{gray}{The lungs appear to have a significant opacity, suggesting possible fluid accumulation or consolidation (\XSolidBrush)}. \textless Reason\textgreater: The central area of the chest X-ray shows increased density, which may indicate the presence of pleural effusion or other pathological processes affecting lung visibility. The clear outlines of the lungs are obscured, which is not typical in a healthy CXR. \textless GPTScore\textgreater: 33.33
\\ \midrule

\texttt{LLaVA-Med} 
& \textless Answer\textgreater: The overexpansion of the right lung with a mediastinal shift to the left could be caused by various factors, such as a buildup of air, fluid, or pressure within the lung cavity. This can be due to \textcolor{gray}{conditions like pneumothorax (collapsed lung), pleural effusion (fluid buildup around the lungs), or a mass causing pressure on the lung (\XSolidBrush)}. It's important to consider the patient's clinical history, symptoms, and additional imaging findings to determine the exact cause of the overexpansion and mediastinal shift. \textless GPTScore\textgreater: 55.56
& \textless Answer\textgreater: The image shows \textcolor{gray}{left lung base fibrocystic disease (\XSolidBrush)}, which is a lung condition characterized by the presence of fibrous and cystic tissue in the lung base. \textless GPTScore\textgreater: 22.22
& \textless Answer\textgreater: In this post-operative chest X-ray (CXR), \textcolor{gray}{the lungs appear to be clear and without any signs of infection (\Checkmark)}. \textless GPTScore\textgreater: 77.78

\\ \midrule

\texttt{LLaVA-Med-\shortname{}} 
& \textless Answer\textgreater: Possible chronic lung disease or \textcolor{gray}{compensatory mechanism (\Checkmark)}. \textless Reason\textgreater: Overexpansion of the lung can be due to chronic lung diseases or compensatory mechanisms, especially \textcolor{orange}{after a left lobectomy (\Checkmark)}. \textless GPTScore\textgreater: 66.67 
&   \textless Answer\textgreater: \textcolor{gray}{Atelectasis (\Checkmark)}. \textless Reason\textgreater: The CXR shows \textcolor{gray}{moderate atelectasis at the left lung base (\Checkmark)}, which is a collapse or incomplete expansion of the lung. \textless GPTScore\textgreater: 77.78
& \textless Answer\textgreater: Clear. \textless Reason\textgreater: The CXR indicates that \textcolor{gray}{the lungs are clear with normal pulmonary vasculature (\Checkmark)}. \textless GPTScore\textgreater: 88.89

\\ \bottomrule
\end{tabular}
}
\vspace{1mm}
\captionof{table}{
Open-ended questions from \shortname{} answered by GPT-4o-mini, LLaVA-Med, and our fine-tuned LLaVA-Med-\shortname{}.}
\label{tab:more_cases_open}  
  \end{minipage}
\end{table*}

\begin{table*}[t]
  \begin{minipage}{0.99\textwidth}
\centering  
\scalebox{0.70}{
\large
\begin{tabular}{l |p{6.5cm}| p{6.5cm} | p{6.5cm} }
\toprule
 \multicolumn{3}{l}{\bf Closed-ended questions from \shortname{}:}  \\
\midrule
& \textbf{CASE I} & \textbf{CASE II} & \textbf{CASE III}\\
&  \includegraphics[height=4.5cm]{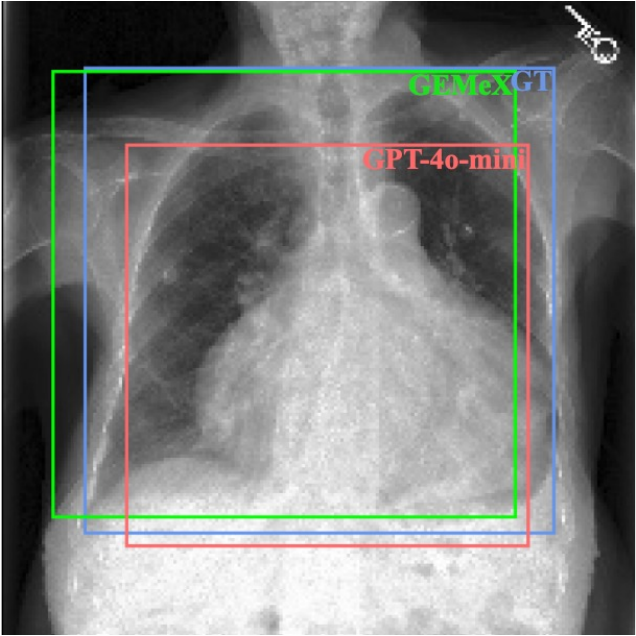} &  \includegraphics[height=4.5cm]{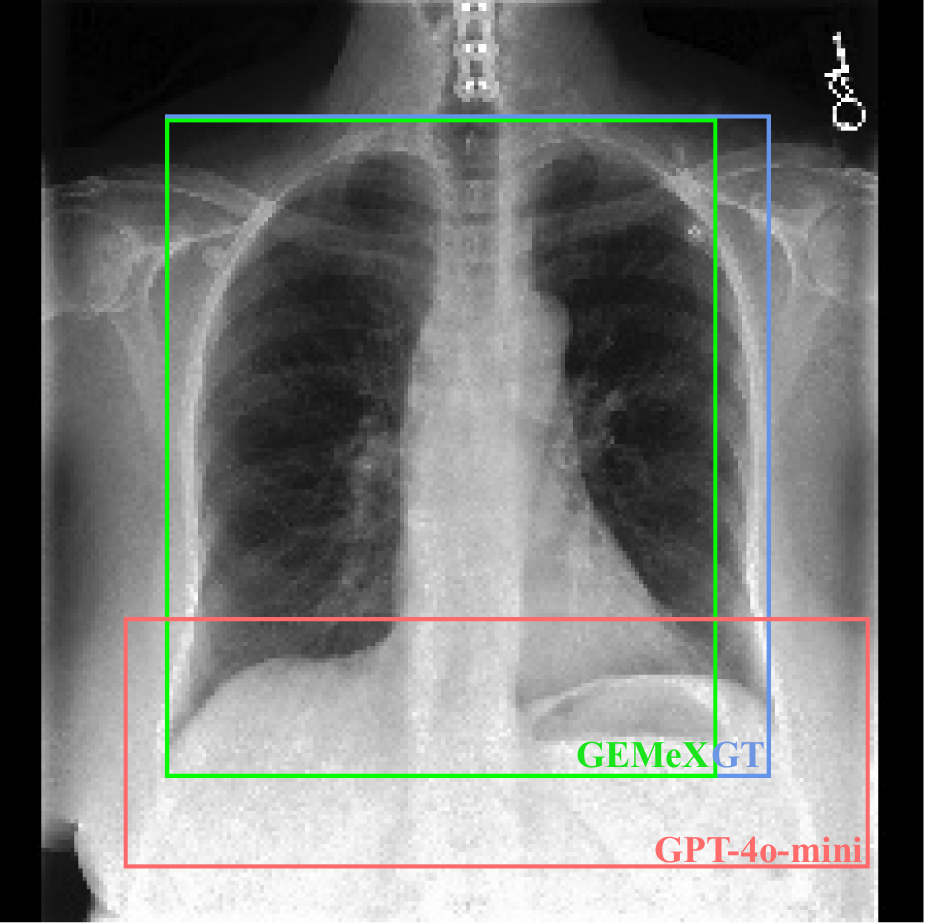} &
\includegraphics[height=4.5cm]{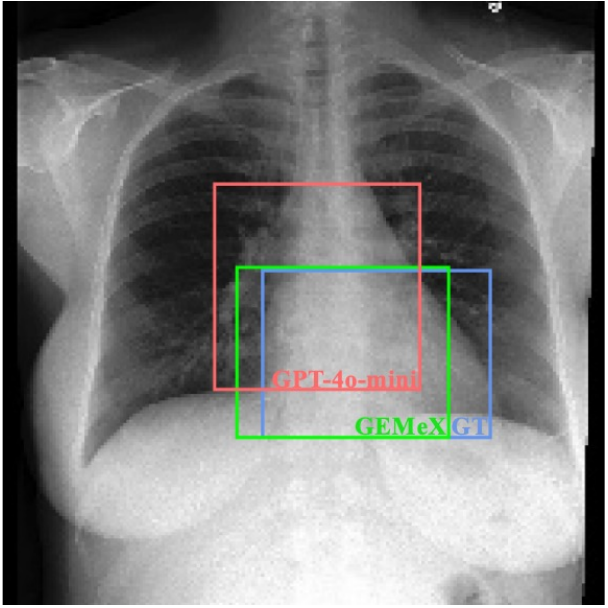}\\
& (Bbox predicted by each model) & (Bbox predicted by each model) &(Bbox predicted by each model)\\
 \midrule
\emph{Question} 
& Are there signs of pneumonia in this CXR?  
& Is there any pleural effusion present in the CXR?
& Is the heart size abnormal in this CXR?
\\

\emph{Ground Truth} 
& \textless Answer\textgreater: No.  \textless Reason\textgreater: The CXR explicitly states that \textcolor{orange}{there is no evidence of pneumonia, indicating clear lung fields}.  
&  \textless Answer\textgreater: No. \textless Reason\textgreater: The CXR \textcolor{orange}{explicitly states that there is no pleural effusion}.
& \textless Answer\textgreater: No. \textless Reason\textgreater: The CXR \textcolor{orange}{indicates that the heart size is within normal limits}.

\\ \midrule
\texttt{GPT-4o-mini} 
&  \textless Answer\textgreater: Yes.   \textless Reason\textgreater: The chest X-ray shows areas of \textcolor{orange}{increased opacity, which may indicate the presence of fluid or consolidation (\XSolidBrush)} typically associated with pneumonia. \textless GPTScore\textgreater: 37.50 
& \textless Answer\textgreater: Yes. \textless Reason\textgreater: The presence of pleural effusion is indicated by the \textcolor{orange}{blunting of the costophrenic angles and the increased opacity in the lower lung fields (\XSolidBrush)}, suggesting fluid accumulation in the pleural space. \textless GPTScore\textgreater: 20.00
&  \textless Answer\textgreater: No. \textless Reason\textgreater: \textcolor{orange}{The heart size appears to be within normal limits (\Checkmark)} based on the overall silhouette and the ratio of the heart to the thoracic cavity. \textless GPTScore\textgreater: 100.00

\\ \midrule

\texttt{LLaVA-Med} 
& \textless Answer\textgreater: No, there are \textcolor{orange}{no signs of pneumonia in this chest X-ray (\Checkmark)}. \textless GPTScore\textgreater: 87.50
& \textless Answer\textgreater: No, \textcolor{orange}{there is no pleural effusion present in the chest X-ray (CXR) (\Checkmark)}. \textless GPTScore\textgreater: 100.00
& \textless Answer\textgreater: Yes, the heart size in this chest X-ray \textcolor{orange}{appears to be abnormally enlarged (\XSolidBrush)}. This is referred to as cardiomegaly. \textless GPTScore\textgreater: 20.00\\

\\ \midrule

\texttt{LLaVA-Med-\shortname{}} 
& \textless Answer\textgreater: No. \textless Reason\textgreater: The CXR indicates \textcolor{orange}{no focal consolidation, which rules out pneumonia (\Checkmark)}. \textless GPTScore\textgreater: 100 
&   \textless Answer\textgreater: No. \textless Reason\textgreater: The CXR \textcolor{orange}{clearly states that there is no pleural effusion (\Checkmark)}. \textless GPTScore\textgreater: 100.00
& \textless Answer\textgreater: No. \textless Reason\textgreater: \textcolor{orange}{The heart is described as normal in size (\Checkmark)}. \textless GPTScore\textgreater: 100.00

\\ \bottomrule
\end{tabular}
}
\vspace{1mm}
\captionof{table}{
Closed-ended questions from \shortname{} answered by GPT-4o-mini, LLaVA-Med, and our fine-tuned LLaVA-Med-\shortname{}.}
\label{tab:more_cases_closed}  
  \end{minipage}
\end{table*}

\begin{table*}[t]
  \begin{minipage}{0.99\textwidth}
\centering  
\vspace{-4mm}
\scalebox{0.70}{
\large
\begin{tabular}{l |p{6.5cm}| p{6.5cm} | p{6.5cm} }
\toprule
 \multicolumn{3}{l}{\bf Single-choice questions from \shortname{}:}  \\
\midrule
& \textbf{CASE I} & \textbf{CASE II} & \textbf{CASE III}\\
&  \includegraphics[height=4.5cm]{images/case_3.pdf} 
&  \includegraphics[height=4.5cm]{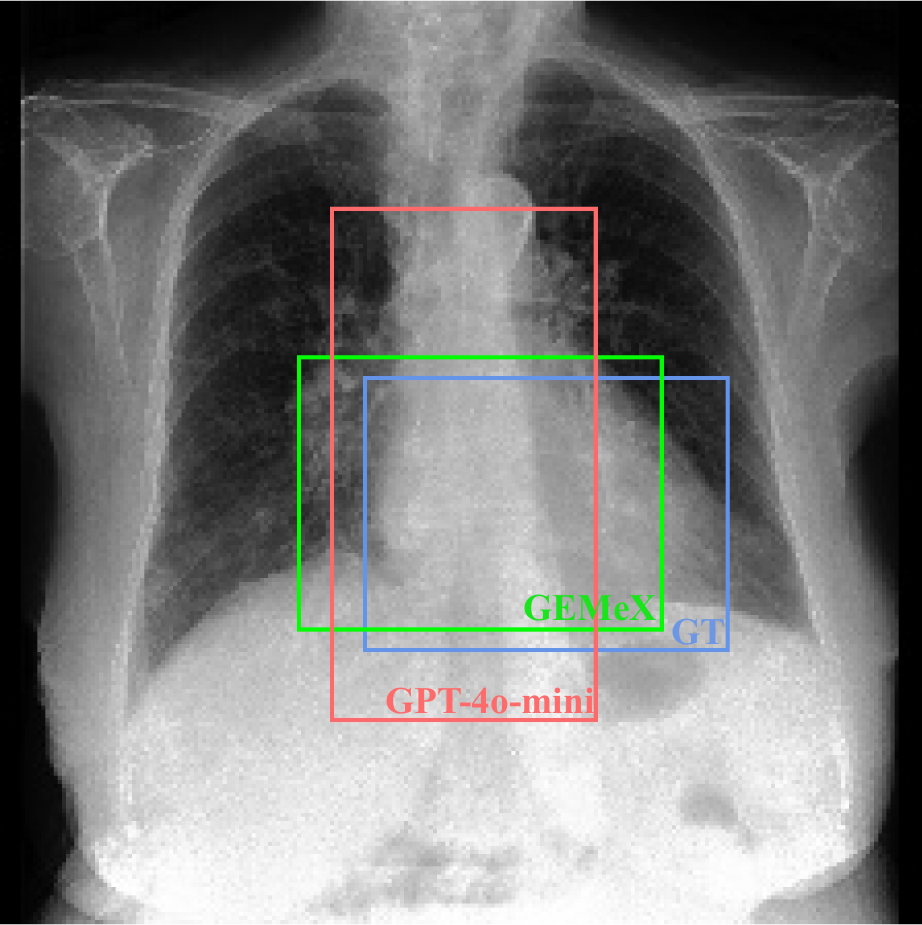} 
&  \includegraphics[height=4.5cm]{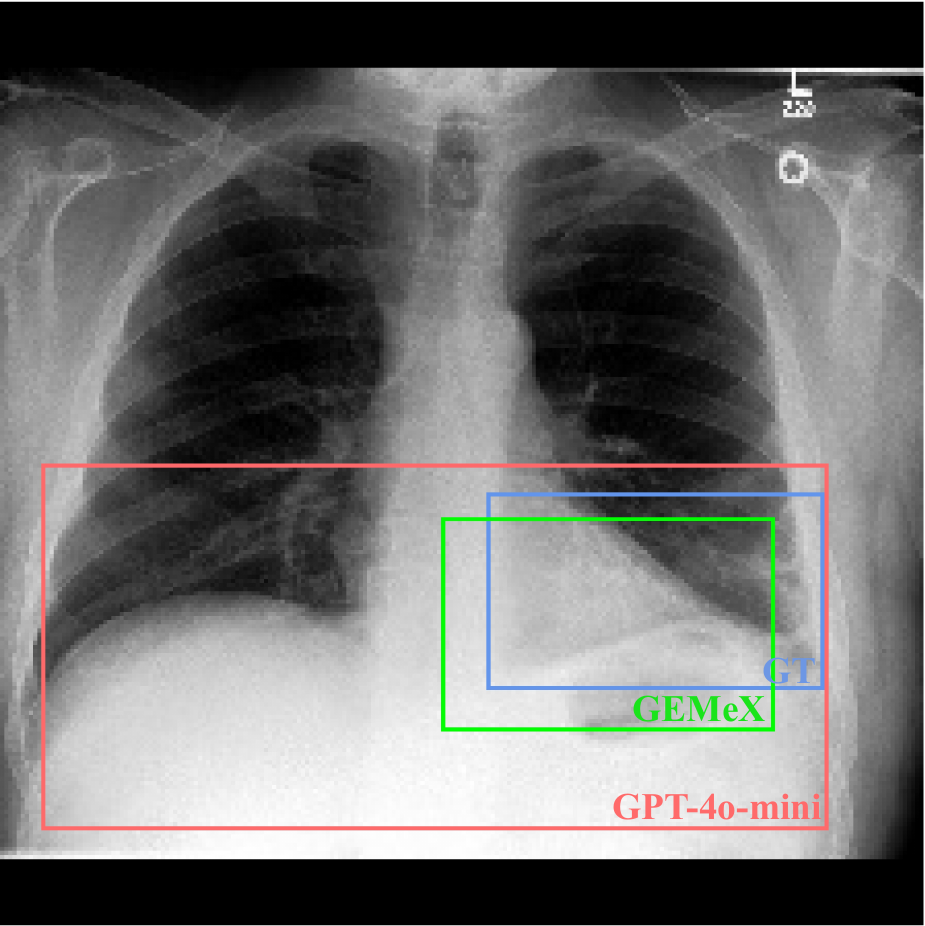}\\
& (Bbox predicted by each model) & (Bbox predicted by each model) &(Bbox predicted by each model)\\
 \midrule
\emph{Question} 
& What is the overall condition of the lungs based on the CXR? Choices: [A: Clear, B: Consolidated, C: Effused, D: Thickened]
& What does the CXR indicate about the heart size? Choices: [A: Normal, B: Mild cardiomegaly, C: Moderate cardiomegaly, D: Severe cardiomegaly]
& Which regions are assessed for pleural effusion in this CXR? Choices: [A: Right lower lung zone, B: Left lower lung zone, C: Cardiac region, D: Bilateral lung]

\\

\emph{Ground Truth} 
& \textless Answer\textgreater: A.  \textless Reason\textgreater: The CXR indicates that \textcolor{orange}{the lungs are clear of consolidation or effusion, although right pleural-based thickening at the base}.
& \textless Answer\textgreater: C. \textless Reason\textgreater: The CXR indicates unchanged \textcolor{orange}{moderate cardiomegaly}.
&  \textless Answer\textgreater: B. \textless Reason\textgreater: \textcolor{orange}{The pleural effusion is specifically noted in the left lower lung zone}.

\\ \midrule

\texttt{GPT-4o-mini} 

&  \textless Answer\textgreater: B. \textless Reason\textgreater: \textcolor{orange}{The presence of increased opacity in the lung fields suggests consolidation, indicating that the lungs are not clear and may have fluid or solid material in the alveoli (\XSolidBrush)}. \textless GPTScore\textgreater: 33.33
& \textless Answer\textgreater: B. \textless Reason\textgreater: The CXR shows \textcolor{orange}{an enlarged heart silhouette, indicating mild cardiomegaly. The heart appears larger than normal but not excessively so (\XSolidBrush)}.  \textless GPTScore\textgreater: 55.56
& \textless Answer\textgreater: D. \textless Reason\textgreater: Pleural effusion can accumulate in the lower lung zones, and both sides of the chest \textcolor{orange}{The bilateral lung regions are crucial for evaluating pleural effusion (\XSolidBrush)}. \textless GPTScore\textgreater: 66.67

\\ \midrule

\texttt{LLaVA-Med} 
& \textless Answer\textgreater: The chest X-ray (CXR) shows \textcolor{orange}{bilateral diffuse interstitial thickening, which means that both lungs have a thickened appearance in their interstitial areas (\XSolidBrush)}. This finding can be associated with various lung conditions, such as infections, inflammation, or fibrosis. It is important to consider the patient's clinical history and symptoms, as well as consult a healthcare professional for a thorough evaluation and proper diagnosis of the underlying cause of these findings.  \textless GPTScore\textgreater: 37.50
& \textless Answer\textgreater: The CXR indicates that there is \textcolor{orange}{mild cardiomegaly, which means that the heart appears slightly enlarged (\XSolidBrush)}. \textless GPTScore\textgreater: 44.44
& \textless Answer\textgreater: In this chest X-ray (CXR), the regions assessed for \textcolor{orange}{pleural effusion include the right lower lung zone, the left lower lung zone, the cardiac region, and the bilateral lung (\XSolidBrush)}. \textless GPTScore\textgreater: 50.00

\\ \midrule

\texttt{LLaVA-Med-\shortname{}} 
& \textless Answer\textgreater: A. \textless Reason\textgreater: The CXR indicates that \textcolor{orange}{the lungs are otherwise clear except for the noted pleural thickening (\Checkmark)}. \textless GPTScore\textgreater: 88.89
&  \textless Answer\textgreater: C. \textless Reason\textgreater: The CXR indicates \textcolor{orange}{moderate cardiomegaly (\Checkmark)}. \textless GPTScore\textgreater: 100.00
& \textless Answer\textgreater: B. \textless Reason\textgreater: The CXR specifically indicates that \textcolor{orange}{trace pleural effusion is difficult to exclude in the left lower lung zone (\Checkmark)}. \textless GPTScore\textgreater: 87.50

\\ \bottomrule
\end{tabular}
}
\vspace{1mm}
\captionof{table}{
Single-choice questions from \shortname{} answered by GPT-4o-mini, LLaVA-Med, and our fine-tuned LLaVA-Med-\shortname{}.}
\label{tab:more_cases_single}  
  \end{minipage}
\end{table*}

\begin{table*}[t]
  \begin{minipage}{0.99\textwidth}
\centering  
\vspace{-4mm}
\scalebox{0.70}{
\large
\begin{tabular}{l |p{6.5cm}| p{6.5cm} | p{6.5cm} }
\toprule
 \multicolumn{3}{l}{\bf Multi-choice questions from \shortname{}:}  \\
\midrule
& \textbf{CASE I} & \textbf{CASE II} & \textbf{CASE III}\\
&  \includegraphics[height=4.5cm]{images/case_2.pdf} 
&  \includegraphics[height=4.5cm]{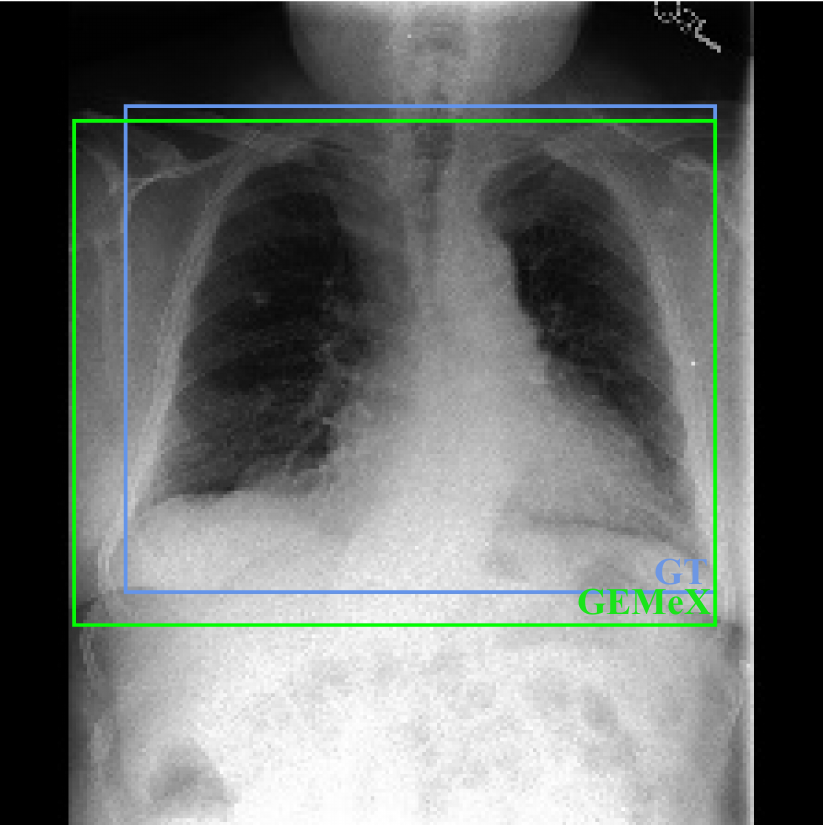} 
&  \includegraphics[height=4.5cm]{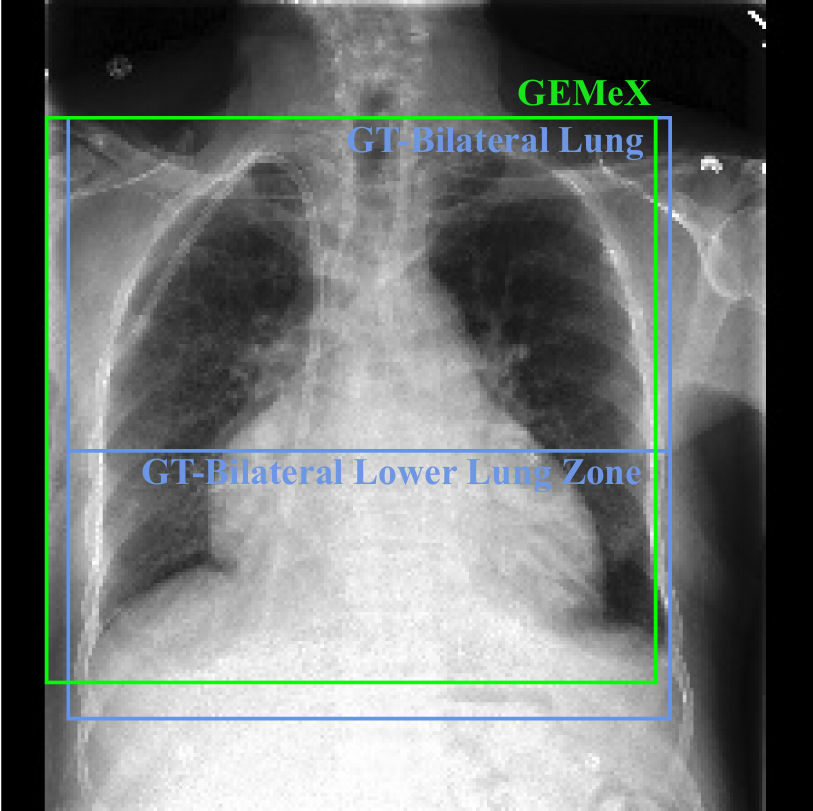}\\
& (Bbox predicted by each model) & (Bbox predicted by each model) &(Bbox predicted by each model)\\
 \midrule
\emph{Question} 
& What regions on the X-ray indicate abnormal findings? Choices: [A: Bilateral lung, B: Cardiac region, C: Abdomen, D: Spine] 
& Which findings indicate abnormal lung conditions? Choices: [A: Mild interstitial abnormality, B: Hyperinflation, C: Focal consolidation, D: No pleural effusion]
& Which observations rule out serious lung conditions in this CXR? [A: No pneumonia, B: No lung nodules, C: No pleural effusions, D: No lymphadenopathy]

\\

\emph{Ground Truth} 
& \textless Answer\textgreater: [A, B, C]. \textless Reason\textgreater: \textcolor{red}{The bilateral lung shows pleural effusion}, \textcolor{gray}{the cardiac region shows enlargement}, and \textcolor{orange}{the abdomen shows the nasogastric tube}. The spine is normal. 
& \textless Answer\textgreater: [A, B]. \textless Reason\textgreater: \textcolor{red}{Mild interstitial abnormality and hyperinflation indicate abnormal lung conditions}. \textcolor{orange}{There is no focal consolidation}. 
&  \textless Answer\textgreater: [A, B, D] \textless Reason\textgreater: \textcolor{red}{The CXR shows no pneumonia, lung nodules, or lymphadenopathy, which rules out serious lung conditions}. \textcolor{orange}{However, there are mild bilateral pleural effusions}.

\\ \midrule

\texttt{GPT-4o-mini} 

&  \textless Answer\textgreater: [A: Bilateral lung]. \textless Reason\textgreater: \textcolor{red}{The bilateral lung region shows potential abnormalities such as opacities or other irregularities that may indicate pathology (\XSolidBrush)}. \textless GPTScore\textgreater: 37.50
&  \textless Answer\textgreater: [A, B, C]. \textless Reason\textgreater: \textcolor{red}{Mild interstitial abnormality, hyperinflation (\Checkmark)}, and \textcolor{orange}{focal consolidation (\XSolidBrush)} are all findings that indicate abnormal lung conditions. \textless GPTScore\textgreater: 66.67
& \textless Answer\textgreater: [A, B, C, D]. \textless Reason\textgreater: \textcolor{red}{The absence of pneumonia, lung nodules, \textcolor{orange}{pleural effusions (\XSolidBrush)}, and lymphadenopathy (\Checkmark)} in the chest X-ray suggests that there are no serious lung conditions present. Each of these findings is associated with significant lung pathology, and their absence helps rule out serious issues. \textless GPTScore\textgreater: 62.50

\\ \midrule

\texttt{LLaVA-Med} 
& \textless Answer\textgreater: The chest X-ray shows abnormal findings in multiple regions. The left side of the image indicates \textcolor{red}{bilateral lung abnormalities, which could be related to lung issues such as infection, inflammation, or other lung conditions (\XSolidBrush)}. The right side of the image shows abnormal findings in the \textcolor{gray}{cardiac region, which may suggest issues with the heart or surrounding structures (\Checkmark)}. \textcolor{orange}{The lower part of the image reveals abnormal findings in the abdomen, which could be related to gastrointestinal or other abdominal issues (\Checkmark)}. Lastly, the spine region also shows abnormal findings, which could be related to spinal issues or other back-related problems. \textless GPTScore\textgreater: 50.00
& \textless Answer\textgreater: The chest x-ray shows findings A, B, and C, which indicate abnormal lung conditions. \textcolor{red}{These findings are likely a combination of interstitial thickening, hyperinflation (\Checkmark)}, and \textcolor{orange}{focal consolidation (\XSolidBrush)}. These abnormalities can be associated with various lung conditions, such as chronic obstructive pulmonary disease (COPD), asthma, or other respiratory disorders. \textless GPTScore\textgreater: 55.56
& \textless Answer\textgreater: The image shows a chest X-ray that demonstrates the absence of certain conditions, which are marked as B on the image. \textcolor{red}{These conditions include pneumonia, lung nodules, lymphadenopathy (\Checkmark)}, and \textcolor{orange}{pleural effusions (\XSolidBrush)}. This means that the chest X-ray does not show any signs of these lung conditions, ruling them out based on the image. \textless GPTScore\textgreater: 62.50

\\ \midrule

\texttt{LLaVA-Med-\shortname{}} 
& \textless Answer\textgreater: [A, B, C]. \textless Reason\textgreater: \textcolor{red}{The bilateral lung shows interstitial opacities (\XSolidBrush)}, \textcolor{gray}{the cardiac region shows enlargement (\Checkmark)} , and \textcolor{orange}{the abdomen shows the feeding tube location (\Checkmark)}.  \textless GPTScore\textgreater: 87.50 
&  \textless Answer\textgreater: [A, B]. \textless Reason\textgreater: The CXR shows \textcolor{red}{mild interstitial abnormality and hyperinflation (\Checkmark)} as abnormal lung conditions. \textless GPTScore\textgreater: 88.89
& \textless Answer\textgreater: [A, B, C, D]. \textless Reason\textgreater: The CXR indicates \textcolor{red}{the absence of pneumonia, lung nodules, \textcolor{orange}{pleural effusions (\XSolidBrush)}, and lymphadenopathy (\Checkmark)}. \textless GPTScore\textgreater: 62.50

\\ \bottomrule
\end{tabular}
}
\vspace{1mm}
\captionof{table}{
Multi-choice questions from \shortname{} answered by GPT-4o-mini, LLaVA-Med, and our fine-tuned LLaVA-Med-\shortname{}.}
\label{tab:more_cases_multi}  
  \end{minipage}
\end{table*}

\end{document}